\documentclass[pdflatex,sn-mathphys-num,iicol]{sn-jnl}

\usepackage{graphicx}%
\usepackage{multicol}
\usepackage{multirow}%
\usepackage{hyperref}
\usepackage{makecell}
\usepackage{amsmath,amssymb,amsfonts}%
\usepackage{amsthm}%
\usepackage{mathrsfs}%
\usepackage[title]{appendix}%
\usepackage{xcolor}%
\usepackage{textcomp}%
\usepackage{manyfoot}%
\usepackage{booktabs}%
\usepackage{algorithm}%
\usepackage{algorithmicx}%
\usepackage{algpseudocode}%
\usepackage{listings}%
\usepackage{dsfont}

\theoremstyle{thmstyleone}%
\theoremstyle{thmstyletwo}%

\theoremstyle{thmstylethree}%

\begin{document}

\title[Article Title]{Toward Robust Medical Fairness: Debiased Dual-Modal Alignment via Text-Guided Attribute-Disentangled Prompt Learning for Vision–Language Models}

\author[1]{\fnm{Yuexuan} \sur{Xia}}\email{xiayuexuan@mail.nwpu.edu.cn}
\equalcont{These authors contributed equally to this work.}

\author[1]{\fnm{Benteng} \sur{Ma}}\email{mabenteng@mail.nwpu.edu.cn}
\equalcont{These authors contributed equally to this work.}

\author[2]{\fnm{Jiang} \sur{He}}\email{ hejiang@huiyihuiying.com}

\author[3]{\fnm{Zhiyong} \sur{Wang}}\email{zhiyong.wang@sydney.edu.au}

\author[4]{\fnm{Qi} \sur{Dou}}\email{qidou@cuhk.edu.hk}

\author*[1]{\fnm{Yong} \sur{Xia}}\email{yxia@nwpu.edu.cn}

\affil[1]{\orgdiv{National Engineering Laboratory for Integrated Aero-Space-Ground-Ocean Big Data Application Technology}, \orgname{Northwestern Polytechnical University}, \orgaddress{\city{Xi’an}, \postcode{710072}, \country{China}}}
\affil[2]{\orgname{Huiying Medical Technology Company Ltd.},\orgaddress{\city{Beijing},\postcode{100192},\country{China}}}
\affil[3]{\orgdiv{The School of Computer Science},\orgname{The University of Sydney},\orgaddress{\city{Sydney},\postcode{NSW 2006},\country{Australia}}
\affil[4]{\orgdiv{Department of Computer Science and Engineering},\orgname{The Chinese University of Hong Kong},\orgaddress{\city{Hong Kong},\postcode{999077},\country{China}}}
}


\abstract{Ensuring fairness across demographic groups in medical diagnosis is essential for equitable healthcare, particularly under distribution shifts caused by variations in imaging equipment and clinical practice. 
Vision–language models (VLMs) exhibit strong generalization, and text prompts encode identity attributes, enabling explicit identification and removal of sensitive directions. However, existing debiasing approaches typically address vision and text modalities independently, leaving residual cross-modal misalignment and fairness gaps.  
To address this challenge, we propose DualFairVL, a multimodal prompt-learning framework that jointly debiases and aligns cross-modal representations. DualFairVL employs a parallel dual-branch architecture that separates sensitive and target attributes, enabling disentangled yet aligned representations across modalities. Approximately orthogonal text anchors are constructed via linear projections, guiding cross-attention mechanisms to produce fused features. A hypernetwork further disentangles attribute-related information and generates instance-aware visual prompts, which encode dual-modal cues for fairness and robustness. Prototype-based regularization is applied in the visual branch to enforce separation of sensitive features and strengthen alignment with textual anchors. 
Extensive experiments on eight medical imaging datasets across four modalities show that DualFairVL achieves state-of-the-art fairness and accuracy under both in- and out-of-distribution settings, outperforming full fine-tuning and parameter-efficient baselines with only 3.6M trainable parameters.
Code will be released upon publication.}




\maketitle

\section{Introduction}\label{sec1}

\begin{figure*}[ht]
\centerline{\includegraphics[width=1\linewidth]{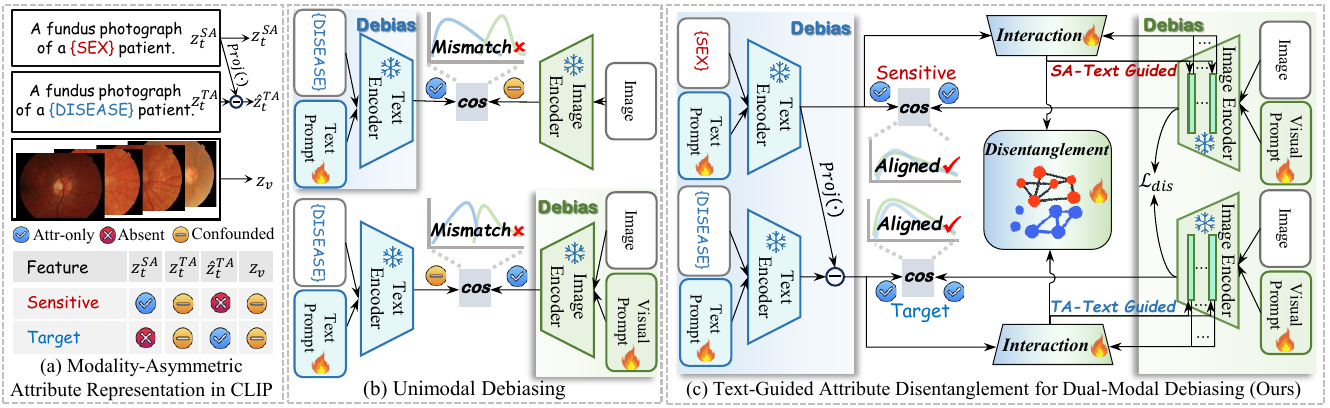}}
\caption{Motivation and overview of DualFairVL. (a) Attribute representation in VLMs (e.g., CLIP). Incorporating the sensitive attribute (SA) into the prompt yields a text embedding $z_t^{\text{SA}}$that exposes the sensitive direction; the target attribute (TA) embedding $z_t^{\text{TA}}$ is debiased via approximately orthogonal projection $Proj(\cdot)$ along $z_t^{\text{SA}}$ to obtain $\hat{z}_{t}^{\text{TA}}$. Conversely, visual features $z_v$ which mix sensitive and target cues are difficult to disentangle. (b) Limits of unimodal debiasing. Debiasing only text or vision breaks cross-modal alignment and leaves residual bias. (c) Our approach. SA/TA text anchors guide cross-attention to produce attribute-specific cross-modal features. A hypernetwork disentangles these features and reconstructs instance-aware visual prompts, which are injected into the image encoder and optimized with $\mathcal{L}_{dis}$. These prompts improve domain robustness and enhance debiased alignment between disentangled visual features and text anchors.}
\label{intro}
\end{figure*}

Fairness in medical diagnosis with deep learning is critical, as biases associated with protected attributes such as race and gender can lead to unequal outcomes and compromise equitable patient care~\cite{zong2022medfair}. While numerous studies have identified algorithmic bias in medical imaging, most assume identical training and testing distributions, overlooking the distribution shifts commonly encountered in clinical practice~\cite{robustfairness}. For instance, differences in retinal imaging modalities across primary care and tertiary hospitals can simultaneously degrade both model accuracy and fairness when models are deployed across institutions. These challenges underscore the urgent need for fairness approaches that remain reliable under domain shifts, thereby ensuring trustworthy deployment of AI in real-world healthcare settings.

Recent benchmarks have started to address domain-shift challenges; however, fairness-aware adaptation of foundation models (FMs) with strong generalization ability remains underexplored~\cite{zong2022medfair}. FMs, including vision models (VMs) and vision–language models (VLMs), leverage large-scale pretraining to acquire transferable representations~\cite{vit,clip}. As these models are increasingly adopted in medical imaging, striking a balance between fairness and utility becomes a central question~\cite{fairmedfm}. Yet, existing strategies seldom exploit the synergy between the strong generalization capabilities of FMs and their potential for debiasing, motivating novel approaches for fairness-aware adaptation. While prior efforts have primarily concentrated on fairness in VMs~\cite{fairvpt,fairdomain,fairtune}, VLMs provide unique advantages through inherent cross-modal alignment, which mitigates bias propagation across modalities. In particular, VLMs have demonstrated robust domain generalization~\cite{clip}, and their ability to encode identity attributes in text enables explicit definition and removal of bias directions during optimization~\cite{debiasvl,biasdirection}.

Medical images present additional challenges: disease-relevant features are distributed across multiple locations and scales and are entangled with sensitive attributes (e.g., sex) and acquisition conditions (e.g., device type). Such entanglement cannot be resolved by a single linear mapping~\cite{orthfairness}. In prompt-based fine-tuning, replacing the disease token in a template such as `a fundus photograph of a [DISEASE] patient' with a sensitive token to form `a fundus photograph of a [SENSITIVE] patient' yields a text embedding that explicitly exposes the sensitive direction, since the only variation arises from the class token~\cite{biasdirection}. Empirical evidence~\cite{debiasvl,prism} confirms that such directions are linearly separable in representation space (see Fig.~\ref{intro}(a)). Because VLMs jointly encode vision and text, prompts can steer the visual pathway toward either sensitive or target-specific directions, enabling debiasing strategies that leverage attribute cues across modalities.

A promising avenue for robust fairness is to exploit VLMs’ ability to capture sensitive attribute directions in the text space, thereby enabling transferable debiasing across domains. However, existing VLM-based debiasing approaches face several limitations. Some require full-model fine-tuning, which is computationally expensive and reduces transferability~\cite{fairclip}. Others debias a single modality—either text~\cite{debiasvl,itc} or vision~\cite{dear,ca,coopood,cfr,mitigating_sc}—leaving residual biases in the complementary modality. This incomplete debiasing disrupts cross-modal alignment and restricts fairness gains, particularly under domain shifts where residual biases are amplified (see Fig.\ref{intro}(b)). Prompt learning offers a compelling alternative: by tuning only a small set of parameters, it provides efficiency and flexibility\cite{goodprompt}. Moreover, attribute-aware text-guided multimodal prompt learning~\cite{vita_clip,vop,maple} can explicitly disentangle attribute information, thereby improving generalization to out-of-distribution (OOD) settings.

Building on these insights, we introduce \textbf{DualFairVL}, a lightweight multimodal debiasing framework based on prompt learning. DualFairVL employs a dual-branch architecture that separately processes sensitive attributes (SA) and target attributes (TA). Approximately orthogonal SA/TA text anchors are constructed to guide cross-modal fusion of attribute-specific information, while instance-aware visual prompts are generated to achieve aligned, debiased representations across modalities. Specifically, a projection in textual space produces stable SA/TA anchors that steer cross-attention to fuse vision–text features. A hypernetwork then disentangles these fused features and reconstructs instance-aware visual prompts encoding attribute-exclusive cues from both modalities, thereby enhancing fairness and domain robustness. These prompts are injected into the visual encoder and further regularized with prototype-based constraints to promote disentanglement and alignment with text anchors. As illustrated in Fig.~\ref{intro}(c), this process yields debiased dual-modal representations that improve both fairness and generalization. 

Our main contributions are as follows:

\begin{itemize}
    \item We propose \textbf{DualFairVL}, a text-guided dual-modal debiasing paradigm that explicitly disentangles sensitive and target attributes, overcoming residual cross-modal biases left by unimodal approaches.

    \item We design a \textbf{dual-branch disentangled representation framework} with instance-aware prompts and prototype-based regularization, producing highly discriminative and debiased target features that generalize across domains.
    
    \item We demonstrate \textbf{state-of-the-art fairness and accuracy} across diverse sensitive attributes and challenging cross-domain scenarios, establishing the effectiveness and robustness of our approach.
    
\end{itemize}

\section{Related Work}
\subsection{Fairness in Medical Imaging}
Fairness in medical imaging aims to ensure equitable diagnostic outcomes across diverse populations by mitigating biases associated with demographic and acquisition factors. Prior work has introduced fairness-oriented datasets~\cite{chexpert,mimiccxr,fitzpatrick17k,kovalyk2022papila,harvardglaucoma,ham10000,fairdomain} and explored debiasing techniques at the pre-processing~\cite{pre_mrfairness,fscl}, in-processing~\cite{orthfairness,groupdro,at_vit}, and post-processing~\cite{postdebiasing,fairprune} stages. However, many of these approaches fail to account for variations in imaging equipment and acquisition protocols across institutions, limiting their robustness to domain shifts. Although several studies have begun to address fairness under distribution shifts~\cite{dgfair_aligning,dgfair_fredom}, most rely on restrictive assumptions about inter-domain relationships. Designing fairness-aware models that generalize reliably across domains remains an open challenge.

\subsection{FMs in Medical Imaging}
FMs trained on large-scale unlabeled data have demonstrated strong generalization across a wide range of tasks. VMs typically leverage masked image modeling (e.g., MedMAE~\cite{medmae}) or contrastive learning (e.g., DINOv2~\cite{dinov2}, MedLVM~\cite{medlvm}), while VLMs employ image–text contrastive learning (e.g., CLIP~\cite{clip}, MedCLIP~\cite{medclip}, BiomedCLIP~\cite{biomedclip}). Despite their adaptability, full fine-tuning of FMs is computationally expensive and prone to overfitting in data-limited medical settings. Prompt learning~\cite{goodprompt} offers a parameter-efficient alternative by optimizing lightweight prompts while freezing model weights. Yet, unimodal prompt learning (either visual~\cite{vpt} or textual~\cite{coop,cocoop}) underutilizes the rich cross-modal information in VLMs. Recent work on multimodal prompt learning~\cite{maple} has shown promise in enhancing cross-modal alignment and generalization, with successful applications in video understanding~\cite{vop,vita_clip} and medical diagnosis~\cite{mcpl}.


\subsection{Fairness in FMs for Medical Imaging}
Fairness research on medical FMs is still at an early stage. Early studies on fairness-aware fine-tuning of VMs demonstrated improvements but suffered from inefficiency and limited generalization. For example, FairDomain~\cite{fairdomain} employs full-parameter tuning with identity attention but risks overfitting, while FairTune~\cite{fairtune} performs validation-guided hyperparameter search at high computational cost. Prompt-based approaches such as FairVPT~\cite{fairvpt} improve feature purity by incorporating fairness into visual prompt tuning~\cite{vpt}; however, their reliance on fixed prompts constrains generalization. 
VLMs inherently offer fairness advantages by exploiting cross-modal alignment to mitigate bias propagation across modalities. They exhibit strong domain generalization~\cite{clip} and can encode identity attributes in text, enabling explicit definition and removal of bias directions~\cite{biasdirection,debiasvl,prism}. Nonetheless, the substantial domain gap between general and medical datasets often results in suboptimal performance of zero-shot fairness methods~\cite{fairerclip} in clinical tasks.

Several debiasing approaches have been proposed for VLMs. FairCLIP~\cite{fairclip}, an early CLIP-based method, improves fairness but requires full fine-tuning, incurring high adaptation costs and reducing generalizability. Parameter-efficient methods such as CPT~\cite{cpt} improve efficiency via prompt tuning but do not explicitly disentangle sensitive features, leaving residual bias. Most disentanglement-based methods remain unimodal: in the text domain, Debias-VL~\cite{debiasvl,prism} uses projection to remove bias, and ITC~\cite{itc} employs adversarial training for unbiased prompts; in the vision domain, DeAR~\cite{dear} and CA~\cite{ca} add fair residual modules, CoOPood~\cite{coopood} disentangles target and sensitive information, and others~\cite{cfr,mitigating_sc} use adversarial training. While these methods reduce bias in one modality, they often introduce residual bias in the other, leading to cross-modal misalignment. JointCLIP~\cite{jointclip} performs symmetric debiasing but ignores domain shifts and the multi-scale entanglement of patient attributes, limiting its effectiveness.
In contrast, our work leverages \textbf{orthogonal text anchors} derived via prompt-based projection to guide \textbf{cross-modal fusion}, addressing both (i) multi-scale visual disentanglement and (ii) alignment across modalities under domain shifts.

\subsection{Positioning of Our Work}
Compared to prior research, our approach is the first to integrate \textbf{text-guided debiasing, dual-branch disentanglement, and cross-modal alignment} into a single lightweight framework. Unlike unimodal methods that leave residual bias, or symmetric debiasing methods that neglect distribution shifts, DualFairVL achieves \textbf{robust and transferable fairness} by jointly disentangling sensitive and target attributes across modalities. This unified design enables our framework to deliver state-of-the-art fairness and accuracy in both in-distribution and cross-domain evaluations.

\begin{figure*}[ht]
\centerline{\includegraphics[width=1.0\linewidth]{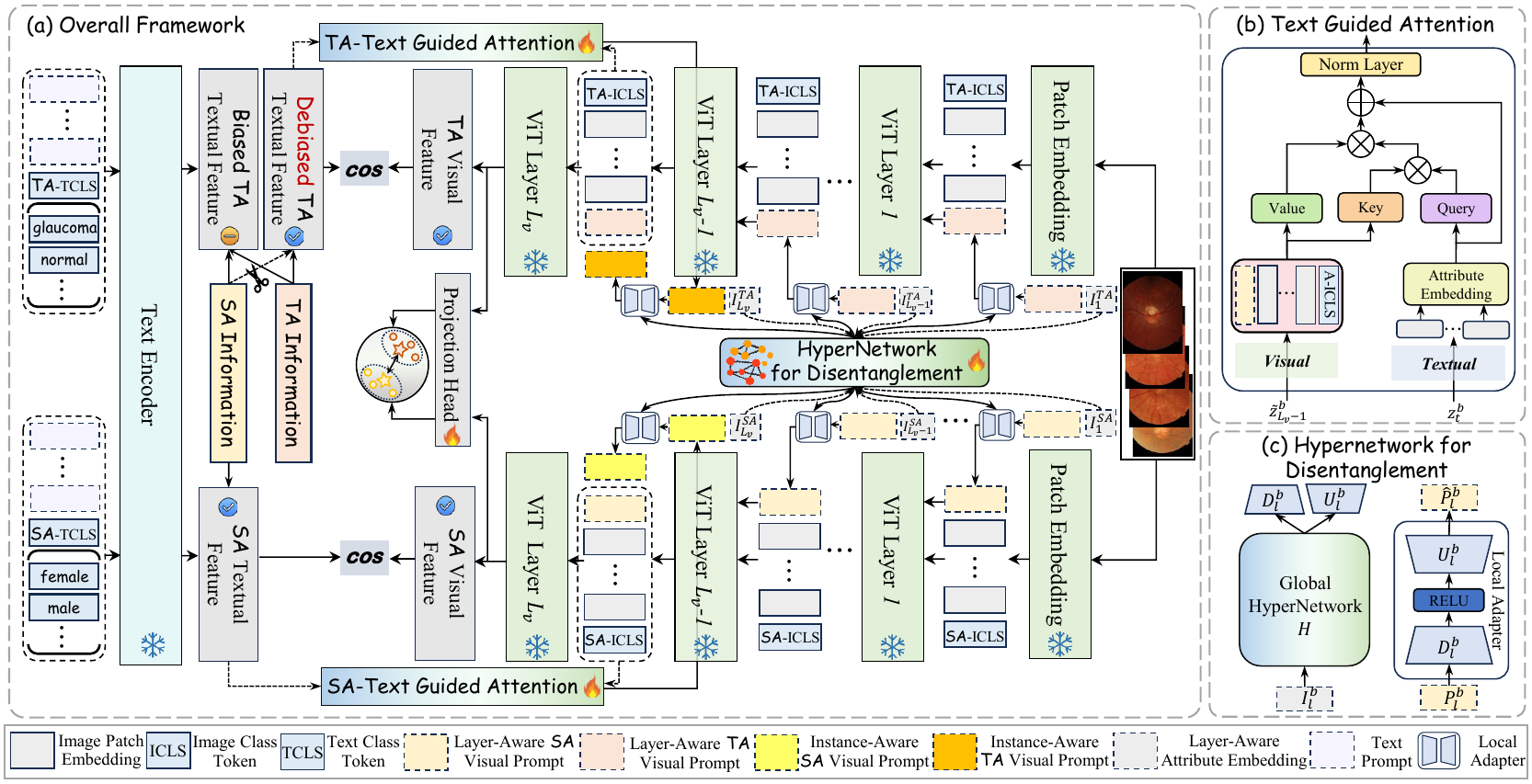}}
\caption{DualFairVL architecture.
(a) Early layers of the Vision Transformer (ViT) are modulated by layer-aware prompts. The text branch produces disentangled SA and TA anchors, which guide cross-modal attention to integrate textual and visual features, producing attribute-aligned and instance-aware prompts that are injected into the final ViT layer. A global hypernetwork processes these joint, layer-wise and modality-mixed representations to disentangle SA and TA information. Prototype-based regularization in the visual branch further enforces separation from the text anchors and improves alignment.
(b) Attribute-specific text anchors steer feature extraction along SA- or TA-specific directions.
(c) The global hypernetwork, conditioned on the branch–layer index, propagates final-layer joint vision–text cues to all layers and generates the corresponding local adapters, thereby enabling attribute-disentangled and layer-specific prompt learning.}
\label{pipeline}
\end{figure*}

\section{Method}
\subsection{Preliminaries}
We consider a dataset $D=\{({{x}_{i}},{{y}_{i}},{{a}_{i}})\}_{i=1}^{N}$, where \( x_i \in X \) denotes the \( i \)-th medical image, \( y_i \in Y \) its diagnostic label, and \( a_i \in A \) a sensitive attribute (e.g., gender, race, or skin type). We adopt a pre-trained VLM, CLIP, consisting of a visual encoder \( \mathcal{V} \) with \( L_v \) Transformer layers \( \{\mathcal{V}_l\}_{l=1}^{L_v} \) and a text encoder \( \mathcal{T} \) with \( L_t \) Transformer layers \( \{\mathcal{T}_l\}_{l=1}^{L_t} \). Prompt tuning is performed on the frozen CLIP model using a learnable parameter set \( P \). Textual prompts are constructed as \( \{p_1^t, p_2^t, \ldots, p_K^t, [\text{CLASS}]\}_{c=1}^C \), where \( p^t = \{p_k^t\}_{k=1}^K \) are learnable text vectors and \( C \) is the number of classes.

Both image and text embeddings are projected into a shared space, yielding features \( z_v,z_t \in \mathbb{R}^{d_{vl}} \). The classifier \( f_{\mathcal{V}, \mathcal{T}, P}: X \to Y \) predicts \( \hat{y}_i \) via temperature-scaled cosine similarity. 
The training objective is the cross-entropy loss

\begin{equation}
    \mathcal{L}_{CE} = -\sum_{i=1}^N \sum_{c=1}^C y_{ic} \log \frac{\exp\left(\langle z_{v,i}, z_{t,c} \rangle / \tau\right)}{\sum_{c'=1}^C \exp\left(\langle z_{v,i}, z_{t,c'} \rangle / \tau\right)},
\end{equation}
where $y_{ic}=\mathds{1}(y_i = c)$ is the one-hot encoding of \( y_i \) and $\tau$ is the temperature parameter.

To achieve fairness, identity attributes must be incorporated, reformulating the predictor as
\begin{equation}
f_{\mathcal{V}, \mathcal{T}, P}: X \times A \to Y,
\end{equation}
so that performance disparities across sensitive groups are minimized. A natural formulation is a dual-branch architecture: a \textbf{target branch} for disease classification and a \textbf{sensitive branch} for identity-related cues. Enforcing disentanglement between branches establishes the foundation for fair representation learning.

\subsection{Overview of DualFairVL}

Fig.~\ref{pipeline} provides an overview of \textbf{DualFairVL}, a multimodal prompt-learning framework designed to achieve fairness that generalizes across domains. The key idea is to disentangle sensitive and target attributes in both modalities through lightweight prompt tuning while maintaining cross-modal alignment. The framework integrates three components:

\noindent \textbf{1. Textual disentanglement:} Sensitive and target attribute directions are identified in text space. Ridge-regularized projection produces orthogonalized text anchors for debiased representations.

\noindent \textbf{2. Cross-modal interaction:} ext anchors guide cross-attention with visual features, generating attribute-specific fused representations. A hypernetwork disentangles these representations and reconstructs instance-aware visual prompts.

\noindent \textbf{3. Visual disentanglement:} Reconstructed prompts are injected into the visual encoder and regularized via a prototype-based loss, enforcing separation of SA and TA features while aligning them with debiased text anchors.

This design yields dual-modal representations that are both \textbf{bias-reduced} and \textbf{domain-generalizable}, with efficiency ensured by training only prompt parameters.

\subsection{Text Anchors Generation}

\subsubsection{Bias Directions in Prompts}
Prior studies demonstrate that text encoders map prompts with sensitive tokens (e.g., ``male") into bias subspaces~\cite{debiasvl,fairerclip}. Due to pretraining imbalance, prompts with disease tokens may also contain spurious SA components. As shown in Fig.~\ref{pipeline}, both branches share prompt parameters $p^t=\{p_k^t\}_{k=1}^{K}$ and the text encoder $\mathcal{T}$. For branch $b\in\{\mathrm{SA},\mathrm{TA}\}$, layerwise textual states are computed as
\begin{equation}
\tilde{W}_l^{\,b}=\mathcal{T}_l(\tilde{W}_{l-1}^{\,b}), \quad 1\le l\le L_t,\quad \tilde{W}_l^{\,b}\in\mathbb{R}^{d_t},
\end{equation}
with the final embedding taken from the [EOS] token at layer $L_t$:
\begin{equation}
z_t^{\,b}=\mathrm{TextProj}\!\left(w_{L_t}^{\,b}\right), \quad z_t^{\,b}\in\mathbb{R}^{d_{vl}}.
\end{equation}
Here, $z_t^{\mathrm{SA}}$ captures sensitive attributes, while $z_t^{\mathrm{TA}}$ encodes disease semantics but may contain leakage.

\subsubsection{Projection for Disentanglement}
To suppress the SA component within the TA feature, ridge regression~\cite{hsr} is used to estimate the SA-aligned part $\hat{S}$ as the projection of $z_t^{\mathrm{TA}}$ onto the direction of $z_t^{\mathrm{SA}}$:
\begin{equation}
\hat{S}
= z_t^{\mathrm{SA}}
\big((z_t^{\mathrm{SA}})^{\top}z_t^{\mathrm{SA}}+\alpha I\big)^{-1}
(z_t^{\mathrm{SA}})^{\top} z_t^{\mathrm{TA}},
\end{equation}
with regularization $\alpha=60$ to mitigate overfitting. The debiased TA embedding is then
\begin{equation}
\hat{z}_t^{\,\mathrm{TA}} = z_t^{\,\mathrm{TA}} - \hat{S}.
\end{equation}
The attribute-specific text embeddings $\{\hat{z}_t^{\,b}\mid b\in\{\mathrm{SA},\mathrm{TA}\}\}$ constitute the text anchors: the TA branch uses the debiased embedding $\hat{z}_t^{\,\mathrm{TA}}$, whereas the SA branch retains the original embedding $\hat{z}_t^{\,\mathrm{SA}}=z_t^{\,\mathrm{SA}}$ for subsequent fairness processing. This operation constitutes a ridge-regularized projection attenuating the SA-aligned component of $z_t^{\,\mathrm{TA}}$; as $\alpha\to 0$ it reduces to orthogonal projection onto the complement of the SA subspace, while a positive $\alpha$ prevents overly aggressive removal of task-relevant semantics.

\subsection{Fairness-Aware Cross-Modal Feature Interaction}  

To ensure that CLIP maintains fairness across diverse domains, the cross-modal fairness-aware interaction module integrates attribute-specific textual cues with instance-level visual information. A hypernetwork processes these fused multimodal features to decouple attribute-related semantics and reconstruct instance-aware visual prompts. This allows each branch to focus on fine-grained, instance-level attribute information, thereby achieving branch-specific cross-modal alignment under domain shift. The text anchor $\hat{z}_{t}^{b}$ serves as the conditional input to this interaction mechanism.

For the visual modality, both branches share a pre-trained visual encoder $\mathcal{V}=\{{{\mathcal{V}}_{l}}\}_{l=1}^{{{L}_{v}}}$, where $l$ is the layer index. Each input image $x$ is divided into $M$ patches and projected into the embedding space ${{E}_{0}}\in {{\mathbb{R}}^{M\times {{d}_{v}}}}$. Within each branch $b$, layer-aware prompts $p_{l}^{v,b}=\{p_{l,k}^{v,b}\}_{k=1}^{K}$ are introduced to capture general branch-specific information, resulting in the visual prompt embedding  
$\tilde{E}_{l}^{b}\in {{\mathbb{R}}^{M\times {{d}_{v}}}}$, where $K$ denotes the prompt token length. The output sequence is then updated via layer-aware prompt tuning for $1\le l\le {{L}_{v}}-1$:

\begin{equation}
   \tilde{z}_{l}^{b}=[cls_{l}^{b},\_,\tilde{E}_{l}^{b}]={{\mathcal{V}}_{l}}([cls_{l-1}^{b},p_{l}^{v,b},\tilde{E}_{l-1}^{b}]),
\end{equation}
where $cls_{l}^{b}$ represents the branch-specific class token. The output sequence $\tilde{z}_{{{L}_{v}}-1}^{b}$ from the $(L_v-1)$-th layer serves as the visual input for interaction.  

\subsubsection{Text Anchor-Guided Attention}
SA/TA text anchors condition cross-attention to fuse textual and visual features and to generate attribute-specific, instance-aware visual prompts (see Fig.~\ref{pipeline}(b)). Following a fully connected attribute embedding projection, the query vector $\mathbf{q}_{c}^{b}\in {{\mathbb{R}}^{{{d}_{v}}}}$ is obtained for each attribute $c$ ($1\le c\le {{C}^{b}}$). The key and value vectors ${{\mathbf{k}}^{b}}\in {{\mathbb{R}}^{n\times {{d}_{v}}}}$ originate from the output sequence $\tilde{z}_{{{L}_{v}}-1}^{b}$ in layer $(L_v-1)$, where $n$ is the length of the sequence. The branch-specific instance-aware visual prompt at layer $L_v$ is computed as  

\begin{equation}
\mathbf{o}_{{{L}_{v}}}^{b,c}=\text{Softmax}\left( \frac{\mathbf{q}_{c}^{b}\mathbf{W}_{q}^{b}{{({{\mathbf{k}}^{b}}\mathbf{W}_{k}^{b})}^{\top }}}{\sqrt{{{d}_{k}}}} \right){{\mathbf{k}}^{b}}\mathbf{W}_{v}^{b},
\end{equation}

\begin{equation}
p_{{{L}_{v}}}^{v,b}=\{\text{LN(}\mathbf{o}_{{{L}_{v}}}^{b,c}+\mathbf{q}_{c}^{b}\text{)  }\!\!|\!\!\text{  1}\le c\le {{C}^{b}}\},
\end{equation}
where $\text{LN}(\cdot)$ denotes the layer normalization function. The projection matrices in the cross-attention mechanism are defined as ${{\mathbf{W}}_{q}}\in {{\mathbb{R}}^{{{d}_{v}}\times {{d}_{k}}}}$, ${{\mathbf{W}}_{k}}\in {{\mathbb{R}}^{{{d}_{v}}\times {{d}_{k}}}}$, and ${{\mathbf{W}}_{v}}\in {{\mathbb{R}}^{{{d}_{v}}\times {{d}_{v}}}}$, with $d_k$ representing the attention head dimension.  

The final input to the last encoder layer is defined as
\begin{equation}
[cls_{{{L}_{v}}}^{b},\_,\tilde{E}_{{{L}_{v}}}^{b}]={{\mathcal{V}}_{{{L}_{v}}}}([cls_{{{L}_{v}}-1}^{b},p_{{{L}_{v}}}^{v,b},\tilde{E}_{{{L}_{v}}-1}^{b}]),
\end{equation}
producing the learnable class token $cl{{s}_{{{L}_{v}}}}$, which is projected through a linear layer into the final visual representation:

\begin{equation}
    z_{v}^{b}=\text{ImageProj}(cls_{{{L}_{v}}}^{b}), \quad z_{v}^{b}\in {{\mathbb{R}}^{{{d}_{vl}}}}.
\end{equation}

\subsubsection{Hypernetwork for Disentanglement}  

We introduce a hypernetwork $H(\cdot)$ to facilitate cross-layer and cross-branch disentanglement of multimodal features through personalized prompt adaptation. This design offers two key advantages: (1) It establishes a global perception mechanism wherein the shared hypernetwork enables all layers in each attribute branch $b$ to simultaneously capture both the fused vision–text representations required for final-layer attribute classification and hierarchical, attribute-specific visual concepts across layers. (2) It provides localized specialization via branch-layer-specific adapters $h_l^b$, which structurally separate parameters between the target and sensitive branches to prevent interference between decoupled features, while also empowering each attribute branch and layer to extract discriminative, task-specific prompt semantics. Specifically, each branch replaces initial prompts $\{p_l^{v,b}\}_{l=1}^{L_v}$ with discriminative variants $\{\hat{p}_l^{v,b}\}_{l=1}^{L_v}$ via hypernetwork-generated adapters:
\begin{equation}
\hat{p}_l^{v,b} = h_l^b(p_l^{v,b}) = U_l^b\left(\text{ReLU}(D_l^b(p_l^{v,b}))\right).
\end{equation}
Layer-aware attribute embeddings $I_l^b \in \mathbb{R}^I$ generate adapter parameters through the global hypernetwork:
\begin{equation}
(U_l^b, D_l^b) = H(I_l^b) = (\mathbf{H}^U, \mathbf{H}^D)I_l^b,
\end{equation}
with global weights $\mathbf{H}^D \in \mathbb{R}^{(d_v\times h)\times I}$, $\mathbf{H}^U \in \mathbb{R}^{(h\times d_v)\times I}$. The embedding $I_l^b$ is constructed as:
\begin{equation}
I_l^b = \begin{cases} 
h_I\left([\mathbf{t}^b; e_l^b(l)]\right), & l < L_v \\ 
h_I(\hat{\mathbf{t}}^b), & l = L_v 
\end{cases},
\end{equation}
where $\mathbf{t}^b \in \mathbb{R}^{I'}$ denotes the attribute branch embedding, $e_l^b(\cdot)$ encodes layer-specific information, and $\hat{\mathbf{t}}^b \in \mathbb{R}^{2I'}$ incorporates instance-aware attributes from the final layer. The fusion MLP $h_I(\cdot)$ consists of two fully-connected layers with ReLU activation. This dual mechanism of global perception and localized specialization enables multi-level disentanglement of target and sensitive features for deep debiasing. In contrast to conventional disentanglement methods~\cite{eccvorthfairness,orthfairness}, which rely on post hoc and rigid separation in a shared feature space, our approach performs dynamic and task-specific customization at the fundamental weight level, thereby preventing information confusion at its source.

\subsection{Visual Disentanglement Learning}  
  
To ensure that the SA and TA branches are disentangled at the visual level, a prototype-based regularization is introduced to explicitly reduce correlations between target and sensitive features. Given a mini-batch ${{\{{{x}_{i}}\}}_{i\in I}}$ sampled from the dataset, a shared projection network $g(\cdot):\mathbb{R}^{d_{vl}} \to \mathbb{R}^{d_p}$ maps visual features $z_{v,i}^b$ to $\tilde{r}_i^b = g(z_{v,i}^b)$ within each branch $b \in \{SA,TA\}$. The normalized embeddings $r_i^b = \tilde{r}_i^b/\|\tilde{r}_i^b\|_2$ within the set $\mathcal{R}^b(i) = \{ r_j^b \mid y_j=y_i, a_j=a_i \}$ reside on a unit hypersphere and are modeled using von Mises-Fisher (vMF) distributions~\cite{vmf}, which are suitable for directional data. The prototype $\mu _{{{y}_{i}},{{a}_{i}}}^{b}\in {{\mathbb{R}}^{{{d}^{p}}}}$ represents the central direction of embeddings in $\mathcal{R}^b(i)$. We first define the compactness loss to promote intra-class clustering:

\begin{equation}
    \mathcal{L}_{com} = -\mathop{\mathbb{E}}\limits_{\substack{b\in B\\ i\in\mathcal I}} \log \frac{\exp\left((r_i^{b})^{\top}\mu_{y_i,a_i}^b/\varphi\right)}{\sum\limits_{b'\in B}\exp\left((r_i^{b})^{\top}\mu_{y_i,a_i}^{b'}/\varphi\right)},
\end{equation}
where $\varphi $ denotes the temperature coefficient. The separability loss aims to further reduce the similarity between prototypes of the two branches:

\begin{equation}
    {{\mathcal{L}}_{sep}}={{\mathbb{E}}_{y\in Y,a\in A}}((\mu {{_{y,a}^{SA}})^{\top }}\mu _{y,a}^{TA}/\varphi ).
\end{equation}

The prototypes ${{\{\mu _{y,a}^{b}\}}_{y\in Y,a\in A}}$ are updated for samples ${{\{{{x}_{i}}\}}_{i\in I}}$ via exponential moving average:

\begin{equation}
\begin{aligned}
\mu_{y,a}^b
&= \operatorname{Normalize}\Big(\beta\,\mu_{y,a}^b\\
&\quad + (1-\beta)\,\mathds{1}\{y_i=y,a_i=a\}\, r_i^b\Big).
\end{aligned}
\end{equation}

The final regularization loss combines both objectives:

\begin{equation}
    \mathcal{L}_{dis} = \mathcal{L}_{com} + \lambda \mathcal{L}_{sep}
\end{equation}
with hyperparameters $\beta=0.5$ and $\lambda=0.1$ showing stable performance. Compared with instance-aware disentanglement~\cite{fairvpt,fscl}, this prototype-based approach robustly captures global category-attribute characteristics while mitigating sample noise effects. The resulting visually disentangled representations are then aligned with the debiased text anchors to complete dual-modal debiasing.

\subsection{Overall Objective Function}
To summarize, our overall objective function is
\begin{equation}
    \mathcal{L}_{total} = \mathcal{L}_{CE}^{TA} + \mathcal{L}_{CE}^{SA} + \delta \mathcal{L}_{dis}.
\end{equation}
Following~\cite{eccvorthfairness}, distinct cross-entropy objectives are assigned to the target-attribute and sensitive-attribute branches, and representation separation is promoted by the disentanglement regularizer $\mathcal{L}_{dis}$ with coefficient $\delta$.

\section{Experiments}

\subsection{Datasets}
We evaluated DualFairVL on eight medical imaging datasets spanning four modalities: optical coherence tomography (OCT), chest X-ray, fundus photography, and dermoscopy. Each dataset provides at least one sensitive attribute (e.g., sex, race, or skin type), enabling both in-distribution (ID) and out-of-distribution (OOD) fairness analyses. Following MedFair~\cite{zong2022medfair}, we binarized task labels and sensitive attributes, and discarded samples with incomplete annotations.

\textbf{ID Evaluation.}
Five datasets were used for ID evaluation: Fitzpatrick17k~\cite{fitzpatrick17k}, PAPILA~\cite{kovalyk2022papila}, Harvard-GF3300~\cite{harvardglaucoma}, HAM10000~\cite{ham10000}, and CheXpert~\cite{chexpert}.

\textit{Fitzpatrick17k}~\cite{fitzpatrick17k}: The three original diagnostic classes were merged into a binary task, where “non-neoplastic” and “benign” form the benign class, and “malignant” remains the malignant class. The sensitive attribute is Fitzpatrick skin type. Data were split into 70\%/10\%/20\% for training/validation/test.

\textit{PAPILA}~\cite{kovalyk2022papila}: The “suspect” category was removed, yielding a binary classification task (glaucomatous vs. non-glaucomatous). Patient-level separation was enforced across training/validation/test (70\%/10\%/20\%). Sex were used as the sensitive attribute.

\textit{Harvard-GF3300}~\cite{harvardglaucoma}: This fairness-oriented dataset contains 2D retinal nerve fiber layer images from 3,300 patients, balanced across racial groups and annotated with age, sex, and race. We conducted binary glaucoma classification, and binarized sex and race following\cite{zong2022medfair}. The official patient-wise split was used.

\textit{HAM10000}~\cite{ham10000}: The seven original categories were consolidated into benign vs. malignant\cite{bin_derm}. Samples lacking sensitive attributes were removed. Sex were used as the sensitive attribute. A 70\%/10\%/20\% split was applied.

\textit{CheXpert}~\cite{chexpert}: Ethnicity annotations were obtained from\cite{processchexpert}. The task is binary classification based on the “No Finding” label. Images without sensitive attributes were discarded, and patient-level splits were maintained across training/validation/test.

\textbf{OOD Evaluation.}
We consider two domain generalization scenarios, differing in acquisition devices, protocols, and sites.

\textit{Dermatology Domain Generalization.}
Three dermoscopy datasets were used: HAM10000, BCN20000~\cite{bcn}, and MSK~\cite{msk}. BCN20000 contains images collected at Hospital Clínic de Barcelona (2010–2016), while MSK is a subset of the ISIC Archive from Memorial Sloan Kettering Cancer Center. All datasets were binarized into benign vs. malignant~\cite{bin_derm}. We evaluated under two settings: (1) \emph{Single-source}: training on HAM10000, testing on BCN20000 and MSK; (2) \emph{Multi-source (leave-one-out)}: training on two datasets, testing on the held-out dataset.

\textit{Chest X-ray Domain Generalization.}
CheXpert serves as the source domain and MIMIC-CXR~\cite{mimiccxr} as the target. Race information for MIMIC-CXR was extracted from MIMIC-IV~\cite{mimiciv} and linked via subject identifiers. Sex and race were treated as sensitive attributes. To simulate a data-limited scenario, we trained on 5\% of CheXpert, and evaluate on 10\% of CheXpert (ID) and 10\% of MIMIC-CXR (OOD).

\subsection{Experimental Setup}

\subsubsection{Implementation Details}
DualFairVL was implemented in PyTorch and trained on four NVIDIA RTX 2080 Ti GPUs. The pipeline consists of two stages: (i) prompt tuning of a CLIP (ViT-B/16) backbone for both the sensitive (SA) and target (TA) branches; and (ii) diagnostic inference using the TA branch on ID and OOD datasets. 
Data augmentation includes random horizontal/vertical flips, zoom, shift, and rotation, followed by resizing to $224\times224$. Following MedFair~\cite{zong2022medfair}, we used AdamW with weight decay $1\mathrm{e}{-4}$, batch size 32, and dataset-specific learning rates between $1\mathrm{e}{-2}$ and $1\mathrm{e}{-3}$. The regularization weight $\delta$ was set to 1. Prompt length was fixed at 4 for both modalities, and the temperature $\tau = 0.1$. Attribute embeddings were initialized with dimension $I'=128$ and projected to $I=32$; the hidden dimension of local adapters was set to $h=24$. Unless otherwise specified, all results were averaged over three independent runs. 

\subsubsection{Metrics}
To jointly assess accuracy and fairness, we report performance using area under the ROC curve (AUC) and fairness using the difference in equalized odds (DEOdds)~\cite{dpd,deodds} and demographic parity difference (DPD)~\cite{deodds}.

DPD quantifies disparities in prediction rates across sensitive groups. It is defined as
\begin{equation}
    \text{DPD} = \tau(A_{\max}) - \tau(A_{\min})
\end{equation}
where \(\tau(A) = E[h(x)\mid a=A]\) is the selection rate of group \(A\), \(A_{\max} = \arg\max_A \tau(A)\) and \(A_{\min} = \arg\min_A \tau(A)\) denote the groups with highest and lowest rates, and \(h(x)\) is the predicted label. $DPD = 0$ indicates perfect parity.

DEOdds evaluates discrepancies in false positive rate (FPR) and true positive rate (TPR) across groups:
\begin{equation}
    \begin{aligned}
    \text{FPR}(A) &= P[h(x) = 1 \mid a = A, Y = 0] \\
    \text{TPR}(A) &= P[h(x) = 1 \mid a = A, Y = 1]
    \end{aligned}
\end{equation}
DEOdds is the maximum inter-group difference in FPR and TPR. Equalized odds is satisfied when $DEOdds = 0$, i.e., all groups share identical FPR and TPR.

\begin{table*}[h]
    \centering
        \caption{In-distribution performance comparison across multiple datasets: HAM10000, PAPILA, Fitzpatrick17k, and Harvard-GF3300. The sensitive attributes evaluated include sex, skin type, and race. Evaluation metrics include AUC [↑, $\%$], DEOdds [↓, $\%$], and DPD [↓, $\%$]. The \textbf{best} and \underline{second-best} results are highlighted.}
        \resizebox{1\textwidth}{!}{%
\begin{tabular}{lccccccccccccccc}
\toprule
\multicolumn{1}{l}{\multirow{2}{*}{Method}} & \multicolumn{3}{c}{HAM10000 (Sex)} & \multicolumn{3}{c}{PAPILA (Sex)} & \multicolumn{3}{c}{Fitzpatrick17k (Skin Type)} & \multicolumn{3}{c}{Harvard-GF3300 (Sex)} & \multicolumn{3}{c}{Harvard-GF3300 (Race)} \\
\cmidrule(l){2-4}\cmidrule(l){5-7}\cmidrule(l){8-10}\cmidrule(l){11-13}\cmidrule(l){14-16}
\multicolumn{1}{c}{}                        & AUC       & DEOdds      & DPD      & AUC      & DEOdds      & DPD     & AUC           & DEOdds          & DPD          & AUC         & DEOdds        & DPD        & AUC         & DEOdds         & DPD        \\
\midrule
\multicolumn{16}{l}{\textit{Vision Prompt Learning}}                                      \\
VPT          & 92.65 & 29.27 & 9.22  & 81.81 & 52.01       & 22.83 & 92.37 & 22.90 & 5.21 & 85.30 & 10.72 & 7.26 & 84.67 & 12.95 & 9.40 \\
VPT+FSCL+    & 92.14 & 23.94 & 10.42 & 80.07 & \underline{11.43} & 10.55 & 90.74 & 20.83 & 8.08 & 84.40 & 12.71 & 6.98 & 85.12 & 11.15 & 7.43 \\
VPT+GroupDRO & 92.83 & 27.53 & 9.22  & 81.55 & 24.59       & 10.18 & 91.60 & 20.27 & 6.02 & 83.36 & 10.32 & 7.32 & 84.10 & 9.47  & 6.67 \\
VPT+AT       & 91.73 & 25.60 & 10.30 & 81.92 & 20.08       & 8.08  & 91.69 & 25.42 & 5.77 & 82.56 & 10.18 & 8.09 & 83.32 & 9.35  & 5.37 \\
FairVPT      & 92.28 & 33.38 & 8.59  & 83.05 & 39.83       & 18.70 & 92.03 & 22.36 & 6.98 & 84.71 & 10.54 & 7.69 & 85.34 & 10.73 & 5.44 \\
\midrule
\multicolumn{16}{l}{\textit{Vision-Language Prompt Learning}}                             \\
CoOp            & 90.45       & 35.64       & 8.75       & 79.46       & 52.85       & 18.70      & 92.56       & 20.05       & 7.52       & 83.78       & 12.21      & 8.06       & 84.09       & 8.47       & 8.33       \\
CoOp+FairCLIP   & 90.60       & 31.21       & 8.91       & 79.12       & 54.44 & 9.25       & 92.61       & \underline{19.55} & 5.70       & 84.61       & 10.93      & 9.26       & 82.74       & 9.62       & 6.00       \\
CoOp+GroupDRO   & 90.77       & 27.32       & 8.39       & 79.98       & 32.64       & 8.81       & 92.14       & 19.79       & 5.52       & 83.73       & 9.99       & 6.86       & 83.58       & 8.17       & 5.91       \\
CoCoOp          & 92.04       & 30.89       & 12.00      & 82.31       & 22.30       & 18.79      & 93.51       & 26.63       & 7.61       & 84.02       & 10.10      & 7.60       & 84.59       & 12.87      & 9.70       \\
CoCoOp+FairCLIP & 91.23       & 24.34       & 8.23       & 81.92       & 20.08       & 8.08       & 93.18       & 20.27       & 6.41       & 83.90       & 13.56      & 7.22       & 83.26       & 10.10      & 7.71       \\
CoCoOp+GroupDRO & 91.36       & 27.97       & 8.84       & 81.19       & 24.11       & 10.02      & 93.01       & 19.59       & 5.56       & 83.83       & 12.86      & 7.15       & 84.57       & 8.81       & 5.53       \\
MaPLe           & 92.19       & 28.94       & 11.96      & 84.61       & 40.90       & 32.18      & 92.63       & 30.84       & 8.68       & 85.20       & 10.11      & 6.40       & \underline{85.63} & 9.04       & 7.25       \\
MaPLe+FairCLIP  & 92.25       & 25.71       & 10.62      & 84.36       & 13.83       & 20.27      & 93.34       & 21.22       & 6.07       & \underline{85.58} & 9.90       & 7.15       & 85.29       & \underline{8.06} & 7.48       \\
MaPLe+GroupDRO  & 92.37       & 27.48       & 11.00      & \underline{84.96} & 19.26       & 16.46      & 93.09       & 21.82       & 5.52       & 85.26       & 9.92       & \underline{6.05} & 84.60       & 8.44       & 6.95       \\
ITC             & 92.23       & 31.01       & 8.96       & 83.99       & 21.80       & 11.79      & \underline{93.82} & 24.76       & 5.97       & 85.19       & 12.27      & 7.64       & 85.06       & 10.20      & 6.64       \\
CPT    & \underline{92.90} & 24.59       & \underline{8.14} & 82.11       & 19.92       & 12.02      & 92.36       & 23.16       & \underline{4.96} & 83.82       & \underline{9.66} & 7.83       & 84.89       & 9.27       & 8.37       \\
CoOPood    & 92.06       & 30.55       & 8.41       & 83.86       & 19.14       & \underline{7.29} & 92.70       & 21.07       & 5.77       & 84.20       & 10.48      & 7.31       & 84.97       & 10.66      & 5.50       \\
VLP+Debias-VL   & 92.70       & \underline{21.42} & 10.70      & 83.73       & 21.51       & 7.49       & 92.17       & 23.80       & 5.17       & 84.91       & 12.36      & 6.93       & 85.29       & 9.48       & \underline{5.20}    \\
\midrule
DualFairVL (Ours)            & \textbf{94.55} & \textbf{16.72} & \textbf{6.22} & \textbf{89.22} & \textbf{2.43} & \textbf{5.04} & \textbf{94.96} & \textbf{18.12} & \textbf{4.56} & \textbf{87.41} & \textbf{8.42} & \textbf{5.54} & \textbf{86.35} & \textbf{7.02} & \textbf{4.18} \\
\bottomrule
        \end{tabular}
    }
\label{tab:in_domain}
\end{table*}

\begin{table*}[htp]
    \centering
        \caption{Comparison of vision and vision-language prompt learning methods under chest X-ray domain generalization (CheXpert → MIMIC-CXR) with sex and race as sensitive attributes.
}
        \resizebox{1.0\textwidth}{!}{%
\begin{tabular}{lccccccccccccc}
\toprule
\multicolumn{1}{l}{\multirow{3}{*}{Method}}  & \multicolumn{6}{c}{CheXpert - MIMIC-CXR (Sex)} & \multicolumn{6}{c}{CheXpert - MIMIC-CXR (Race)} &\multirow{3}{*}{\makecell[c]{Trainable\\Param.}} \\
\cmidrule(l){2-4}\cmidrule(l){5-7}\cmidrule(l){8-10}\cmidrule(l){11-13}
                        & \multicolumn{3}{c}{Source} & \multicolumn{3}{c}{Target} & \multicolumn{3}{c}{Source} & \multicolumn{3}{c}{Target} & \\
                        & AUC    & DEOdds   & DPD   & AUC    & DEOdds    & DPD   & AUC    & DEOdds   & DPD   & AUC     & DEOdds   & DPD   &   \\
\midrule
\multicolumn{14}{l}{\textit{Vision Prompt Learning}}                \\
VPT          & 79.74 & 27.28 & 3.27 & 75.02 & 9.07       & 4.38 & 79.89 & 23.08       & 4.03 & 75.03 & 9.17 & 6.05 & 0.1M                  \\
VPT+FSCL+    & 78.15 & 19.53 & 4.07 & 73.63 & \underline{6.28} & 3.23 & 78.69 & 17.92       & 3.56 & 74.49 & 7.81 & 4.73 & 0.1M                  \\
VPT+GroupDRO & 79.53 & 22.55 & 2.59 & 74.60 & 8.88       & 3.98 & 78.97 & \underline{15.10} & 4.20 & 75.12 & 9.16 & 5.00 & 0.1M                  \\
VPT+AT       & 77.66 & 19.88 & 3.34 & 74.17 & 7.44       & 3.42 & 77.72 & 20.52       & 2.89 & 73.65 & 6.05 & 3.63 & 0.1M                  \\
FairVPT      & 79.08 & 20.95 & 3.08 & 75.69 & 7.34       & 3.37 & 79.06 & 15.58       & 3.36 & 74.33 & 7.64 & 5.29 & 0.5M                 \\
\midrule
\multicolumn{14}{l}{\textit{Vision-Language Prompt Learning}}       \\
CoOp            & 78.06          & 23.56          & 3.74          & 75.15          & 11.13         & 5.41          & 78.15          & 19.53          & 4.07          & 74.74          & 8.45          & 6.00          & 8K \\
CoOp+FairCLIP   & 79.85          & 22.90          & 3.50          & 73.89          & 8.85    & 3.59          & 77.04          & 18.20          & 3.62          & 73.60          & 5.97          & 4.21          & 8K \\
CoOp+GroupDRO   & 79.71          & \underline{19.29}    & \underline{1.37}    & 74.54          & 6.83          & 3.83          & 79.03          & 17.28    & 3.36          & 74.14          & 6.21          & 3.08          & 8K \\
CoCoOp          & 78.50          & 20.15          & 4.32          & 74.65          & 9.06          & 5.12          & 78.80          & 25.67          & 3.78          & 74.40          & 9.12          & 5.39          & 42K \\
CoCoOp+FairCLIP & 77.72          & 20.52          & 2.89          & 74.44          & 9.91          & 4.34          & 78.95          & 16.92          & 2.65          & 74.15          & 6.16          & 3.82          & 42K \\
CoCoOp+GroupDRO & 79.28          & 22.11          & 1.75          & 75.14          & 8.57          & 3.95          & 79.32          & 15.78          & \underline{2.32}    & 75.07          & 5.95          & 3.30          & 42K \\
MaPLe           & 80.05          & 30.54          & 3.36          & 76.08          & 10.25         & 5.22          & \underline{80.51}    & 20.10          & 2.87          & 76.31          & 11.44         & 6.70          & 3.6M \\
MaPLe+FairCLIP  & 79.28          & 21.55          & 3.93          & 75.94          & 7.06          & \underline{3.05}    & 79.76          & 23.27          & 3.76          & 75.89          & 6.32          & 3.55          & 3.6M \\
MaPLe+GroupDRO  & 80.18          & 24.82          & 2.51          & \underline{76.42}    & 9.32          & 4.25          & 80.11          & 34.82          & 3.47          & 76.03          & 5.15          & \underline{2.55}    & 3.6M \\
ITC             & 80.11          & 34.82          & 3.47          & 76.29          & 12.62         & 6.46          & 80.37          & 19.34          & 2.43          & 76.05          & 10.77         & 8.13          & 30.7M \\
CPT    & 79.94          & 19.55          & 2.18          & 74.65          & 9.06          & 5.12          & 78.85          & 17.98          & 3.99          & 74.45          & 8.58          & 6.09          & 0.1M \\
CoOPood    & 80.18          & 20.64          & 1.98          & 76.13          & 7.21          & 3.17          & 79.23          & 16.15          & 2.52          & \underline{76.49}    & \underline{4.86}    & 3.04          & 1.0M \\
VLP+Debias-VL   & \underline{80.62}    & 20.17          & 1.65          & 75.90          & 12.79         & 5.48          & 79.18          & 18.82          & 4.38          & 76.43          & 5.32          & 3.12          & 37K \\
\midrule
DualFairVL (Ours)            & \textbf{83.00} & \textbf{18.28} & \textbf{0.88} & \textbf{78.40} & \textbf{4.57} & \textbf{2.03} & \textbf{82.07} & \textbf{14.14} & \textbf{1.74} & \textbf{78.96} & \textbf{3.36} & \textbf{1.79}          & 3.6M \\
\bottomrule
\end{tabular}
    }
\label{tab:chexpert_out_domain}
\end{table*}

\begin{table*}[htp]
    \centering
        \caption{Comparison under dermatology domain generalization with sex as the sensitive attribute, evaluating single-source (training on HAM10000) and multi-source (leave-one-out) adaptation across three dermoscopy datasets.}
        \resizebox{1.0\textwidth}{!}{%
\begin{tabular}{lccccccccccccccc}
\toprule
\multicolumn{1}{l}{\multirow{3}{*}{Method}}  & \multicolumn{6}{c}{Single-Source (Source: HAM10000)} & \multicolumn{9}{c}{Multi-Source (Leave-one-out)}  \\
\cmidrule(l){2-7}\cmidrule(l){8-16}
                        & \multicolumn{3}{c}{BCN20000} & \multicolumn{3}{c}{MSK}& \multicolumn{3}{c}{HAM10000} & \multicolumn{3}{c}{BCN20000} & \multicolumn{3}{c}{MSK} \\
                        & AUC    & DEOdds   & DPD   & AUC    & DEOdds    & DPD   & AUC    & DEOdds   & DPD   & AUC     & DEOdds   & DPD   & AUC     & DEOdds   & DPD \\
\midrule
\multicolumn{16}{l}{\textit{Vision Prompt Learning}}                \\
VPT          & 68.83 & 19.34 & 4.38       & 73.52 & 15.21 & 4.31 & 77.12 & 25.46 & 12.35 & 73.12 & 15.86 & 3.75 & 77.54 & 12.91 & 3.23       \\
VPT+FSCL+    & 67.75 & 17.28 & 3.52       & 72.26 & 13.79 & 3.29 & 78.58 & 24.97 & 10.03 & 72.86 & 14.52 & 3.62 & 76.65 & 11.12 & 3.44       \\
VPT+GroupDRO & 68.61 & 16.59 &  \underline{3.23} & 74.45 & 13.45 & 5.71 & 79.41 & 26.64 & 11.65 & 72.91 & 13.27 & 4.08 & 79.68 & 10.29 & 2.74       \\
VPT+AT       & 69.85 & 18.27 & 4.02       & 73.53 & 13.76 & 3.65 & 77.15 & 22.67 & 9.43  & 71.54 & 14.68 & 3.54 & 77.12 & 11.8  & 2.26       \\
FairVPT      & 69.93 & 19.54 & 4.16       & 75.55 & 12.48 & 3.01 & 79.45 & 18.48 & 8.31  & 74.25 & 15.64 & 4.42 & 78.48 & 9.73  &  \underline{1.13}\\
\midrule
\multicolumn{16}{l}{\textit{Vision-Language Prompt Learning}}       \\
CoOp            & 69.54          & 15.22          & 4.07          & 77.07          & 15.84        & 4.28          & 78.17          & 24.52          & 11.59         & 75.36          & 14.89         & 3.25          & 79.81       & 12.98         & 3.28          \\
CoOp+FairCLIP   & 70.24          & 14.88          & 3.75          & 76.28          & 13.64        & 3.97          & 77.26          & 22.85          & 9.8           & 75.9           & 13.38         & 2.96          & 78.5        & 10.26         & 2.21          \\
CoOp+GroupDRO   & 71.54          & 16.85          &  \underline{3.82}    & 78.34          & 12.94        & 3.16          & 79.49          & 19.89          & 9.54          & 75.24          & 12.69         & 2.86          & 80.67       & 11.42         & 2.28          \\
CoCoOp          & 70.87          & 16.69          & 5.21          & 78.23          & 13.79        & 4.44          & 79.61          & 20.28          & 10.86         & 74.65          & 12.53         & 2.77          & 81.59       & 12.15         & 4.78          \\
CoCoOp+FairCLIP & 71.14          & 15.03          & 5.42          & 78.6           & 11.04        &  \underline{2.79}    & 78.67          & 19.25          & 11.93         & 74.32          &  \underline{12.11}   & 2.83          & 80.52       & 9.4           &  \underline{3.67}    \\
CoCoOp+GroupDRO & 71.56          & 17.86          & 3.91          & 77.19          & 10.11        & 4.02          & 79.34          & 18.74          & 13.55         & 75.87          & 13.86         & 3.01          & 79.26       & 8.94          & 2.89          \\
MaPLe           & 72.51          & 19.43          & 5.07          & 80.61          & 12.9         & 5.18          & 80.64          & 26.68          &  \underline{8.64}    & 76.52          & 17.59         & 5.24          & 82.57       & 10.25         & 2.94          \\
MaPLe+FairCLIP  & 73.45          & 15.33          & 3.62          & 81.36          & 10.83        & 4.27          & 81.21          & 25.3           & 9.41          & 76.23          & 13.68         & 2.84          &  \underline{83.63} & 8.54          & 3.92          \\
MaPLe+GroupDRO  & 74.56          &  \underline{14.35}    & 3.98          & 81.02          & 9.41         & 3.74          & 82.75          & 22.89          & 11.56         &  \underline{77.04}    & 12.54         & 3.48          & 81.21       & 7.02          & 1.85          \\
ITC             & 71.54          & 18.99          & 4.57          &  \underline{81.82}    & 13.89        & 4.94          & 82.47          & 24.84          & 13.26         & 76.09          & 15.49         & 4.19          & 83.08       & 13.62         & 4.52          \\
CPT             & 73.04          & 17.72          & 3.94          & 80.86          & 11.52        & 3.78          & 81.78          & 19.05          & 9.35          & 75.84          & 14.27         & 3.19          & 82.95       & 10.07         & 3.56          \\
CoOPood         & 72.23          & 16.59          & 3.54          & 81.28          &  \underline{9.31}   & 4.87          &  \underline{83.29}    &  \underline{18.41}    & 8.84          & 76.81          & 14.85         & 3.51          & 82.78       & 8.56          & 2.77          \\
VLP+Debias-VL   &  \underline{74.62}    & 15.89          & 3.29          & 80.75          & 12.08        & 3.62          & 82.08          & 20.87          & 10.29         & 75.86          & 13.54         &  \underline{2.75}    & 83.23       &  \underline{7.23}    & 2.04          \\
\midrule
DualFairVL (Ours)            & \textbf{77.32} & \textbf{12.74} & \textbf{2.11} & \textbf{84.21} & \textbf{5.6} & \textbf{1.26} & \textbf{85.84} & \textbf{16.52} & \textbf{6.01} & \textbf{80.22} & \textbf{9.45} & \textbf{1.24} & \textbf{86} & \textbf{4.03} & \textbf{0.97}\\
\bottomrule
\end{tabular}
    }
\label{tab:skin_out_domain}
\end{table*}

\subsection{Comparison with State-of-the-Art Methods}  
\subsubsection{In-Distribution Results}  
Table~\ref{tab:in_domain} reports comparisons with fairness-aware vision and vision-language prompt tuning methods on HAM10000, PAPILA, Fitzpatrick17K, and Harvard-GF3300 under gender, skin type, and race attributes.

\textbf{Vision Prompt Learning Methods.}
We benchmarked five vision prompt tuning methods, with VPT~\cite{vpt} as the baseline. Fairness-enhanced variants include FSCL+\cite{fscl}, AT\cite{at_vit}, GroupDRO~\cite{groupdro}, and FairVPT~\cite{fairvpt}. These methods produced marginal AUC gains over VPT, and in many cases degraded predictive accuracy. Fairness outcomes were inconsistent, e.g., FairVPT reduced DPD by 0.63\% but increased DEOdds by 4.11\% on HAM10000. By contrast, DualFairVL consistently improved both accuracy and fairness across datasets, indicating more effective in-distribution debiasing.

\textbf{Vision-Language Prompt Learning Methods.} 
Thirteen methods were evaluated, including standard vision-language prompt tuning (CoOp~\cite{coop}, CoCoOp~\cite{cocoop}, MaPLe~\cite{maple}), and fairness-regularized methods (FairCLIP~\cite{fairclip}, GroupDRO~\cite{groupdro}), and fairness-aware prompt tuning approaches (ITC~\cite{itc}, CPT~\cite{cpt}, CoOPood~\cite{coopood}, and VLP~\cite{maple}+Debias-VL~\cite{debiasvl}). Multi-modal prompt tuning generally achieved higher AUC than unimodal methods, but fairness gains were inconsistent. Many approaches improved one fairness metric while degrading another. For instance, VLP+Debias-VL increased DPD by 1.74\%, indicating limited debiasing capability. In contrast, DualFairVL effectively mitigated data bias while enhancing dual-modal alignment. On PAPILA, it outperformed prior methods with a 4.26\% AUC gain and reductions of 9\% and 2.25\% in DEOdds and DPD, respectively.

\begin{figure*}[ht]
\centerline{\includegraphics[width=1\linewidth]{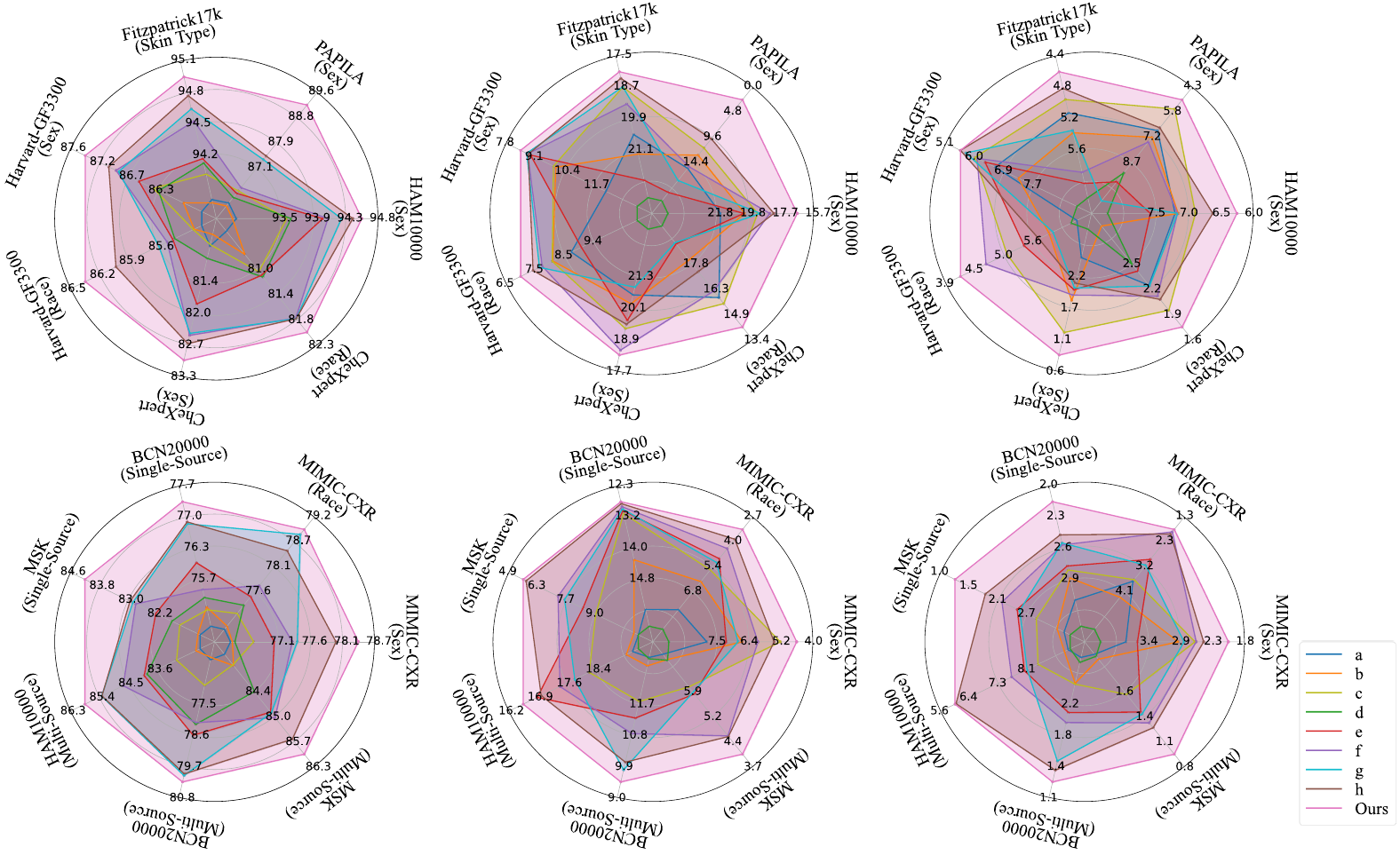}}
\caption{Ablation results of DualFairVL under in-distribution (top) and out-of-distribution (bottom) settings across multiple datasets. Radar charts (left to right) represent AUC, DEOdds, and DPD metrics.}
\label{radar_chart}
\end{figure*}

\begin{table}[h]
\centering
\caption{Ablation on components and trainable parameters: visual debiasing ($\mathcal{L}_{dis}$), cross-modal debiasing with hypernetwork $H(\cdot)$ and text-guided cross-attention $Attn(\cdot)$, and textual debiasing via $Proj(\cdot)$.}
\begin{tabular}{l c cc c c}    
\toprule
\multirow{2}{*}{Model} & Vision & \multicolumn{2}{c}{Cross-Modal} & Text & \multirow{2}{*}{Param.} \\
\cmidrule(l){2-2}\cmidrule(l){3-4}\cmidrule(l){5-5}
& $\mathcal{L}_{dis}$ & $H(\cdot)$ & $Attn(\cdot)$ & $Proj(\cdot)$ & \\
\midrule
a    &                &            &            & \checkmark & 39K  \\
b    & \checkmark     &            &            &            & 0.2M \\
c    & \checkmark     &            &            & \checkmark & 0.2M \\
d    &                &            & \checkmark &            & 1.1M \\
e    & \checkmark     &            & \checkmark &            & 2.3M \\
f    & \checkmark     &            & \checkmark & \checkmark & 2.3M \\
g    &                & \checkmark & \checkmark &            & 3.6M \\
h    & \checkmark     & \checkmark & \checkmark &            & 3.6M \\
Ours & \checkmark     & \checkmark & \checkmark & \checkmark & 3.6M \\
\bottomrule
\end{tabular}
\label{tab:ablation}
\end{table}

\subsubsection{Domain Generalization Results} 
Tables~\ref{tab:chexpert_out_domain} and \ref{tab:skin_out_domain} summarize the results of chest X-ray and dermatology domain generalization under gender and race attributes.

\textbf{Vision Prompt Learning Methods.} 
The five vision prompt learning methods assessed in the in-distribution setting were also evaluated here. Although VPT often achieved the highest AUC, fairness-oriented variants typically reduced DEOdds and DPD at the cost of accuracy. This trade-off was particularly evident under domain shifts: in chest X-ray generalization, the best target-domain AUC did not exceed 76\%, while in dermatology generalization the single-source setting yielded AUC values below 70\% on BCN20000. Even in the multi-source leave-one-out setup, performance remained under 75\%. These results suggest that vision-only prompt learning primarily captures superficial image features and is highly sensitive to training-domain biases, leading to limited robustness and unstable fairness under distribution shift.


\textbf{Vision-Language Prompt Learning.} 
The thirteen vision–language prompt learning methods used for in-distribution evaluation were also compared in this setting. Notably, VLP+Debias-VL applies Debias-VL only to the text branch, resulting in asymmetric debiasing. Overall, vision–language methods exhibited more stable AUC values on target domains compared to vision-only approaches, highlighting their stronger generalization capability. Multi-modal prompt tuning (e.g., MaPLe and its variants) improved AUC through enhanced cross-modal alignment but yielded only marginal fairness benefits relative to unimodal methods such as CoOp and CoCoOp. Techniques incorporating fairness constraints (e.g., FairCLIP, GroupDRO, CoOPood) further reduced fairness disparities, yet often sacrificed accuracy because they lacked explicit disentanglement of sensitive attributes and imposed rigid fairness objectives. In dermatology domain generalization, the multi-source setup consistently outperformed the single-source setting, indicating that diverse training domains facilitate stronger cross-domain generalization.
Debiasing only the visual modality~\cite{coopood} or only the textual modality~\cite{itc,debiasvl} disrupted cross-modal consistency, leading to limited improvements in fairness and accuracy. In contrast, \textbf{DualFairVL} jointly debiases both modalities while preserving cross-modal alignment, thereby achieving consistent gains in both fairness and predictive performance across domain shifts.

\begin{table*}[h]
    \centering
        \caption{Comparison with full-parameter fine-tuning, fairness-aware full-parameter fine-tuning on foundation models (FMs), and parameter-efficient fine-tuning methods on vision-language models (VLMs) across multiple datasets: HAM10000, PAPILA, Fitzpatrick17k, and Harvard-GF3300.}
        \resizebox{1\textwidth}{!}{%
\begin{tabular}{lccccccccccccccc}
\toprule
\multicolumn{1}{l}{\multirow{2}{*}{Method}} & \multicolumn{3}{c}{HAM10000 (Sex)} & \multicolumn{3}{c}{PAPILA (Sex)} & \multicolumn{3}{c}{Fitzpatrick17k (Skin Type)} & \multicolumn{3}{c}{Harvard-GF3300 (Sex)} & \multicolumn{3}{c}{Harvard-GF3300 (Race)} \\
\cmidrule(l){2-4}\cmidrule(l){5-7}\cmidrule(l){8-10}\cmidrule(l){11-13}\cmidrule(l){14-16}
\multicolumn{1}{c}{}                        & AUC       & DEOdds      & DPD      & AUC      & DEOdds      & DPD     & AUC           & DEOdds          & DPD          & AUC         & DEOdds        & DPD        & AUC         & DEOdds         & DPD        \\
\midrule
\multicolumn{16}{l}{\textit{Full-Parameter Fine-Tuning on Pretrained FMs}}                                      \\
ViT        & 87.35       & 27.97       & 9.10        & 84.13       & 24.46       & 16.46      & 90.89       & 24.42       & 7.13       & 84.09       & 12.68      & 9.58       & 83.05       & 10.83      & 9.48       \\
DINOv2     & 86.22       & 28.57       & 10.56      & 83.78       & 27.62       & 15.24      & 86.07       & 25.74       & 7.48       & 85.01       & 10.51      & 8.74       & 82.19       & 11.08      & 12.11      \\
MedMAE     & 89.62       & 25.73       & 10.42      & 82.24       & 24.4        & 11.07      & 91.26       & 27.09       & 7.81       & 85.08       & 11.29      & 8.97       & 82.23       & 10.81      & 8.54       \\
CLIP       & 91.84       & 21.35       & 7.49       & 86.48       & 17.70        & 8.01       & 92.26       & 20.18       & 5.58       & 85.71       & 9.85       & 7.35       & 84.35       & 9.04       & 6.67       \\
BiomedCLIP & 92.13       & 20.24       & \underline{7.17} & \underline{87.46} & 17.61       & 7.19       & 92.54       & \underline{19.08} & \underline{4.92} & \underline{86.12} & 9.79       & 6.84       & 84.47       & 8.61       & 5.96       \\
\midrule
\multicolumn{16}{l}{\textit{Fairness-Aware Full-Parameter Fine-Tuning on Pretrained FMs}}                                      \\
FairDomain & 90.51       & 23.94       & 8.85       & 84.21       & 23.02       & 10.85      & 91.51       & 24.05       & 6.93       & 85.27       & 10.48      & 8.43       & 83.14       & 10.24      & 6.99       \\
FairCLIP   & \underline{92.83} & \underline{17.43} & 7.39       & 87.37       & 15.41       & 7.25       & 92.91       & 19.45       & 5.03       & 85.82        & 9.83       & 6.91       & 84.41       & 8.32       & \underline{5.85} \\
\midrule
\multicolumn{16}{l}{\textit{Other Fairness-Aware Parameter-Efficient Fine-Tuning on Pretrained VLMs}}                                \\
CA         & 91.26       & 23.22       & 8.04       & 85.9        & 20.55       & 9.24       & 92.09       & 20.94       & 6.14       & 85.40        & 10.38      & 7.82       & 83.70        & 9.66       & 6.80        \\
CFR        & 91.36       & 22.65       & 8.30        & 85.86       & 21.54       & 8.07       & 91.95       & 21.85       & 6.21       & 85.59       & 9.97       & 7.79       & 83.48       & 9.71       & 6.78       \\
JointCLIP  & 92.14       & 20.04       & 7.18       & 87.06       & \underline{13.57} & \underline{6.81} & \underline{92.97} & 19.91       & 5.51       & 86.03       & \underline{9.52} & \underline{6.82} & \underline{84.54} & \underline{8.15} & 5.91          \\
\midrule
DualFairVL (Ours)            & \textbf{94.55} & \textbf{16.72} & \textbf{6.22} & \textbf{89.22} & \textbf{2.43} & \textbf{5.04} & \textbf{94.96} & \textbf{18.12} & \textbf{4.56} & \textbf{87.41} & \textbf{8.42} & \textbf{5.54} & \textbf{86.35} & \textbf{7.02} & \textbf{4.18} \\
\bottomrule
        \end{tabular}
    }
\label{tab:in_domain_other}
\end{table*}

\begin{table*}[h]
    \centering
        \caption{Comparison with full-parameter fine-tuning, fairness-aware full-parameter fine-tuning on FMs, and parameter-efficient fine-tuning methods on VLMs under chest X-ray domain generalization (CheXpert → MIMIC-CXR) with sex and race as sensitive attributes.
}
        \resizebox{1.0\textwidth}{!}{%
\begin{tabular}{lccccccccccccc}
\toprule
\multicolumn{1}{l}{\multirow{3}{*}{Method}}  & \multicolumn{6}{c}{CheXpert - MIMIC-CXR (Sex)} & \multicolumn{6}{c}{CheXpert - MIMIC-CXR (Race)} &\multirow{3}{*}{\makecell[c]{Trainable\\Param.}} \\
\cmidrule(l){2-4}\cmidrule(l){5-7}\cmidrule(l){8-10}\cmidrule(l){11-13}
                        & \multicolumn{3}{c}{Source} & \multicolumn{3}{c}{Target} & \multicolumn{3}{c}{Source} & \multicolumn{3}{c}{Target} & \\
                        & AUC    & DEOdds   & DPD   & AUC    & DEOdds    & DPD   & AUC    & DEOdds   & DPD   & AUC     & DEOdds   & DPD   &   \\
\midrule
\multicolumn{14}{l}{\textit{Full-Parameter Fine-Tuning on Pretrained FMs}}                \\
ViT        & 79.68       & 28.96       & 3.07       & 74.15       & 12.73      & 6.97       & 79.17       & 21.67      & 3.82       & 74.46       & 13.89      & 10.06      & 86M  \\
DINOv2     & 80.63       & 27.46       & 3.75       & 73.70        & 11.76      & 7.16        & 79.02       & 25.99      & 4.65       & 74.14       & 15.58      & 10.71      & 87M  \\
MedMAE     & 79.72       & 24.68       & 3.83       & 73.49       & 10.14      & 6.16       & 80.14       & 25.43      & 3.53       & 75.03       & 15.29      & 11.25      & 86M  \\
CLIP       & 81.21       & 21.11       & 2.34       & 75.76       & 8.02       & 4.17       & 80.87       & 17.47      & 2.56       & 76.17       & 9.92       & 6.39       & 150M \\
BiomedCLIP & 81.18       & 20.47       & \underline{1.97} & \underline{77.28} & 7.68       & 3.95       & 81.02          & \underline{16.40} & 2.45       & 76.44       & \underline{6.99} & \underline{4.92} & 196M \\
\midrule
\multicolumn{14}{l}{\textit{Fairness-Aware Full-Parameter Fine-Tuning on Pretrained FMs}}                \\
FairDomain & 80.66       & 23.51       & 2.95       & 74.84       & 9.50        & 4.71       & 80.26       & 20.53      & 2.67       & 75.08       & 13.2       & 8.05       & 93M  \\
FairCLIP   & \underline{81.56} & \underline{20.19} & 2.01       & 76.13       & 7.85       & \underline{3.25} & \underline{81.74} & 17.33      & \underline{2.11} & \underline{76.69} & 9.35       & 5.20        & 150M \\
\midrule
\multicolumn{14}{l}{\textit{Other Fairness-Aware Parameter-Efficient Fine-Tuning on Pretrained VLMs}}                \\
CA         & 80.67       & 22.76       & 2.34       & 75.42       & 9.10        & 4.32       & 80.56       & 17.83      & 2.65       & 75.58       & 10.75      & 7.69       & 0.2M \\
CFR        & 81.03       & 21.65       & 2.78       & 75.13       & 8.35       & 4.28       & 80.69       & 18.01      & 2.57       & 75.92       & 11.43      & 7.94       & 0.4M \\
JointCLIP  & 81.24       & 20.85       & 2.06       & 75.93       & \underline{6.54} & 3.67       & 81.26       & 16.74      & 2.38       & 76.55       & 8.24       & 5.97       & 0.5M \\
\midrule
DualFairVL (Ours)            & \textbf{83.00} & \textbf{18.28} & \textbf{0.88} & \textbf{78.40} & \textbf{4.57} & \textbf{2.03} & \textbf{82.07} & \textbf{14.14} & \textbf{1.74} & \textbf{78.96} & \textbf{3.36} & \textbf{1.79}          & 3.6M \\
\bottomrule
\end{tabular}
    }
\label{tab:chexpert_out_domain_other}
\end{table*}

\begin{table*}[h]
    \centering
        \caption{Comparison with full-parameter fine-tuning, fairness-aware full-parameter fine-tuning on FMs, and parameter-efficient fine-tuning methods on VLMs under dermatology domain generalization with sex as the sensitive attribute, evaluating single-source (training on HAM10000) and multi-source (leave-one-out) adaptation across three dermoscopy datasets.}
        \resizebox{1.0\textwidth}{!}{%
\begin{tabular}{lccccccccccccccc}
\toprule
\multicolumn{1}{l}{\multirow{3}{*}{Method}}  & \multicolumn{6}{c}{Single-Source (Source: HAM10000)} & \multicolumn{9}{c}{Multi-Source (Leave-one-out)}  \\
\cmidrule(l){2-7}\cmidrule(l){8-16}
                        & \multicolumn{3}{c}{BCN20000} & \multicolumn{3}{c}{MSK}& \multicolumn{3}{c}{HAM10000} & \multicolumn{3}{c}{BCN20000} & \multicolumn{3}{c}{MSK} \\
                        & AUC    & DEOdds   & DPD   & AUC    & DEOdds    & DPD   & AUC    & DEOdds   & DPD   & AUC     & DEOdds   & DPD   & AUC     & DEOdds   & DPD \\
\midrule
\multicolumn{16}{l}{\textit{Full-Parameter Fine-Tuning on Pretrained FMs}}                \\
ViT        & 72.22       & 16.76       & 4.58       & 79.00          & 13.27     & 5.88       & 80.25       & 20.32       & 10.74      & 73.51       & 13.81       & 4.78       & 82.17       & 8.85       & 4.13       \\
DINOv2     & 72.51       & 17.8        & 5.31       & 78.70        & 12.63     & 4.23       & 79.94       & 20.43       & 11.19      & 74.11       & 14.57       & 4.40        & 81.61       & 8.24       & 3.40        \\
MedMAE     & 72.42       & 16.24       & 4.53       & 79.40        & 12.76     & 5.47       & 79.98       & 19.96       & 11.12      & 75.34       & 15.17       & 3.95       & 80.98       & 8.23       & 3.17       \\
CLIP       & 73.99       & 14.40        & 4.12       & 80.65       & 10.37     & 4.17       & 82.06       & 19.39       & 8.93       & 75.73       & 12.78       & 3.35       & 82.62       & 7.38       & 2.51       \\
BiomedCLIP & 74.5        & 14.32       & 4.07       & 80.78       & 8.24      & 3.63       & \underline{83.32} & 19.26       & \underline{7.81} & 77.50        & 11.58       & 2.80        & \underline{84.53} & 7.00          & 2.48       \\
\midrule
\multicolumn{16}{l}{\textit{Fairness-Aware Full-Parameter Fine-Tuning on Pretrained FMs}}                \\
FairDomain & 72.86       & 15.02       & 4.51       & 79.81       & 11.36     & 4.22       & 80.98       & 19.88       & 10.73      & 75.47       & 13.60        & 3.86       & 82.22       & 8.04       & 3.09       \\
FairCLIP   & 74.11       & 14.30        & 3.95       & 80.82       & 8.60       & 4.07       & 82.37       & \underline{18.26} & 8.80        & 76.13       & \underline{12.65} & 2.90        & 83.55       & 7.17       & \underline{1.86} \\
\midrule
\multicolumn{16}{l}{\textit{Other Fairness-Aware Parameter-Efficient Fine-Tuning on Pretrained VLMs}}                \\
CA         & 73.21       & 14.92       & 4.23       & 80.56       & 10.59     & 4.22       & 81.71       & 19.63       & 9.63       & 75.64       & 12.82       & 3.64       & 82.57       & 7.95       & 3.07       \\
CFR        & 73.85       & 14.96       & 4.33       & 79.90        & 10.42     & 4.19       & 81.61       & 19.47       & 10.52      & 75.66       & 12.85       & 3.38       & 82.48       & 7.93       & 2.92       \\
JointCLIP  & \underline{75.08} & \underline{14.06} & \underline{3.89} & \underline{81.46} & \underline{7.40} & \underline{3.21} & 83.02       & 18.51       & 8.23       & \underline{77.85} & 12.69       & \underline{2.01} & 83.03       & \underline{6.28} & 2.15      \\
\midrule
DualFairVL (Ours)            & \textbf{77.32} & \textbf{12.74} & \textbf{2.11} & \textbf{84.21} & \textbf{5.60} & \textbf{1.26} & \textbf{85.84} & \textbf{16.52} & \textbf{6.01} & \textbf{80.22} & \textbf{9.45} & \textbf{1.24} & \textbf{86.00} & \textbf{4.03} & \textbf{0.97}\\
\bottomrule
\end{tabular}
    }
\label{tab:skin_out_domain_other}
\end{table*}

\subsection{Ablation Study}

Table~\ref{tab:ablation} summarizes the module configurations and parameter counts of the ablation models, while Fig.~\ref{radar_chart} visualizes their effects under both in-distribution and out-of-distribution evaluations.

\subsubsection{Effect of Textual Debiasing}
Applying textual debiasing via $Proj(\cdot)$ alone (setting \textit{a}) resulted in suboptimal performance. Comparisons such as \textit{b} vs. \textit{c}, \textit{e} vs. \textit{f}, and \textit{h} vs. \textit{Ours} reveal that integrating textual debiasing with either visual or cross-modal strategies consistently improves both accuracy and fairness. These results suggest that text-only debiasing is insufficient: residual visual biases remain and disrupt cross-modal alignment. Comprehensive fairness and robust generalization therefore require joint debiasing across both modalities.

\subsubsection{Effect of Cross-Modal Interaction}
The text-guided cross-attention mechanism $Attn(\cdot)$ enhanced AUC in both in-distribution and out-of-distribution settings (e.g., \textit{b} vs. \textit{e}, \textit{c} vs. \textit{f}), highlighting the value of instance-aware visual prompt tuning. However, fairness metrics showed inconsistent improvement, likely due to residual modality-specific biases in the fused features. When combined with the hypernetwork $H(\cdot)$ for disentanglement (e.g., \textit{d} vs. \textit{g}), the model more effectively separated attribute-related information, yielding consistent gains in both fairness and predictive performance.

\subsubsection{Effect of Visual Debiasing}
Visual debiasing was implemented through the prototype-based regularizer $\mathcal{L}_{dis}$. Incorporating $\mathcal{L}_{dis}$ (e.g., \textit{a} vs. \textit{c}, \textit{g} vs. \textit{h}) improved both performance and fairness by enforcing separation of sensitive and target features. Nevertheless, improvements were limited when textual bias was not simultaneously addressed, as instance-aware prompts could still embed biased information. Only when both modalities were jointly debiased (\textit{Ours}) did the model achieve consistent improvements across all metrics, underscoring the importance of dual-modal debiasing for robust cross-domain fairness and alignment.

\begin{table}[h]
  \centering
  \caption{Analysis of impact of prompt length and sharing strategies, presenting averaged scores across in-distribution (ID) and out-of-distribution (OOD) settings.}
  \setlength\tabcolsep{0.1mm}{
\begin{tabular}{ccccccc}
\toprule
\multirow{2}{*}{Model}              & \multicolumn{3}{c}{ID}     
& \multicolumn{3}{c}{OOD}                                                 \\
\cmidrule(l){2-4}\cmidrule(l){5-7}
                               & \multicolumn{1}{c}{AUC} & \multicolumn{1}{c}{DEOdds} & \multicolumn{1}{c}{DPD} & \multicolumn{1}{c}{AUC} & \multicolumn{1}{c}{DEOdds} & \multicolumn{1}{c}{DPD} \\
\midrule
\multicolumn{7}{l}{\textit{Analysis of Prompt Length}}                             \\  
2  & 87.93          & 12.53          & 4.64          & 80.29          & 11.53         & 4.19          \\
4  & \textbf{88.22} & \textbf{12.16} & \textbf{4.02} & \textbf{81.56} & \textbf{8.04} & \textbf{2.20} \\
8  & 87.94          & 12.82          & 4.35          & 80.79          & 9.63          & 2.88          \\
16 & 87.61          & 13.24          & 4.70           & 80.56          & 9.25          & 3.24         \\
\midrule
\multicolumn{7}{l}{\textit{Analysis of Prompt Sharing Approaches}}                             \\
Layer-Aware     & 87.86          & 12.54          & 4.83          & 79.54          & 5.25          & 2.41          \\
Instance-Aware  & 87.95          & 12.81          & 4.51          & 80.73          & 4.54          & 5.08          \\
Combined (Ours) & \textbf{88.22} & \textbf{12.16} & \textbf{4.02} & \textbf{81.56} & \textbf{8.04} & \textbf{2.20}\\
\bottomrule
\end{tabular}
    }
\label{tab:prompt_analysis}
\end{table}

\begin{figure*}[h]
\centerline{\includegraphics[width=1.0\linewidth]{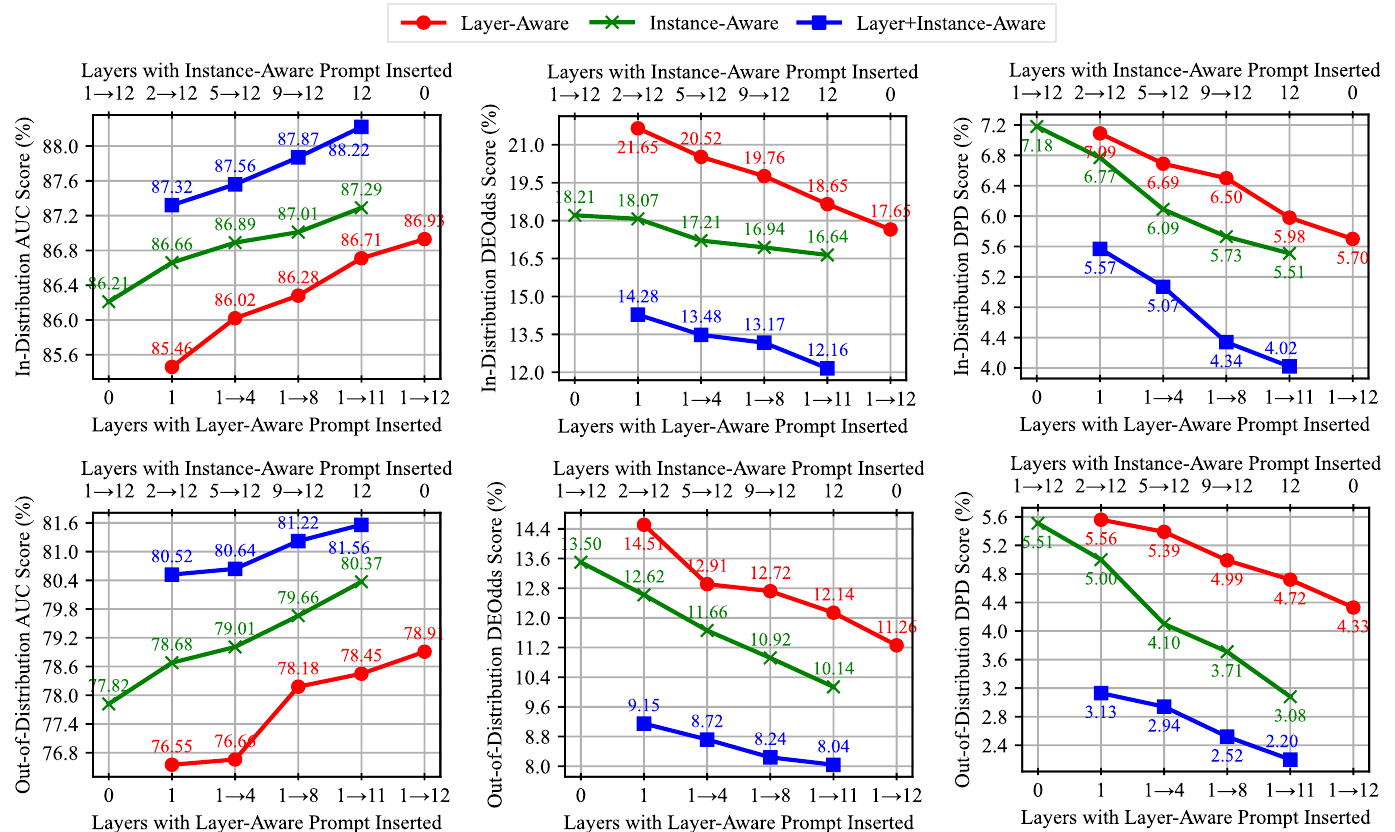}}
\caption{Analysis of influence of prompt types and depth. The bottom and top x-axes indicate the insertion layers for layer-aware prompts and instance-aware prompts, respectively.}
\label{prompt_depth}
\end{figure*}

\section{Discussion}

\subsection{Comparison with Other Fine-Tuning Methods}
We compared DualFairVL with full-parameter fine-tuning on five pretrained backbones, including three vision models (VMs: ViT~\cite{vit}, DINOv2~\cite{dinov2}, MedMAE~\cite{medmae}) and two vision–language models (VLMs: CLIP~\cite{clip}, BiomedCLIP~\cite{biomedclip}). In addition, we evaluated two fairness-aware full fine-tuning methods (FairDomain~\cite{fairdomain}, FairCLIP~\cite{fairclip}) and three parameter-efficient fairness-oriented approaches for VLMs (CA~\cite{ca}, CFR~\cite{cfr}, JointCLIP~\cite{jointclip}). As shown in Tables~\ref{tab:in_domain_other}, \ref{tab:chexpert_out_domain_other}, and \ref{tab:skin_out_domain_other}, VLMs consistently outperform VMs in cross-domain generalization. FairDomain and FairCLIP improve fairness under OOD conditions but offer limited benefit in in-distribution cases. While CA and CFR are computationally efficient, they deliver inferior fairness and accuracy under domain shifts. JointCLIP attempts dual-modal debiasing but relies on symmetric linear mapping without cross-modal fusion, limiting its ability to address acquisition-induced shifts and multi-scale visual attribute entanglement. In contrast, DualFairVL achieves the best overall trade-off, delivering superior fairness and AUC with only 3.6M trainable parameters.

\begin{figure}[h]
\centerline{\includegraphics[width=1.0\linewidth]{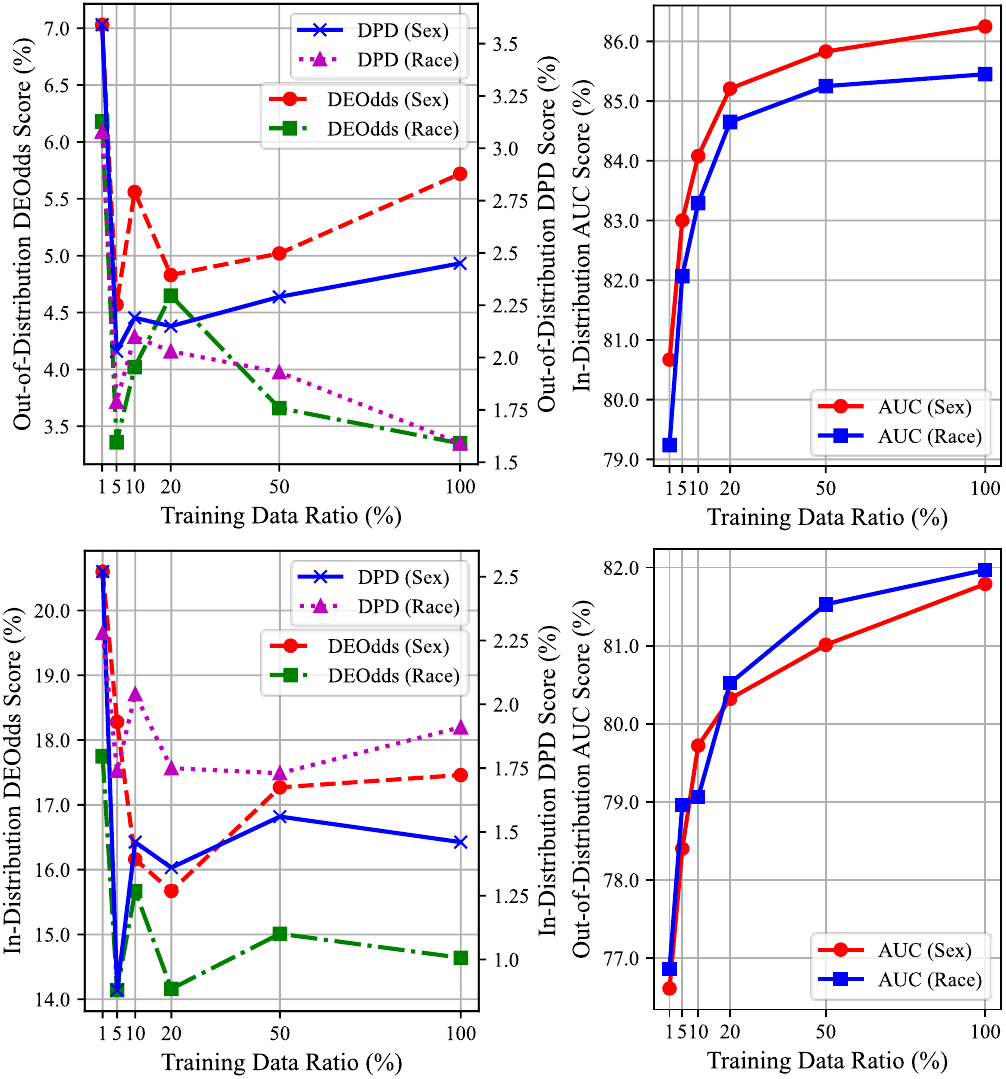}}
\caption{Robustness analysis with varying training data ratios. DualFairVL was trained on CheXpert and tested on CheXpert (in-distribution) and MIMIC-CXR (out-of-distribution).}
\label{training_ratio}
\end{figure}

\begin{figure*}[h]
\centerline{\includegraphics[width=1\linewidth]{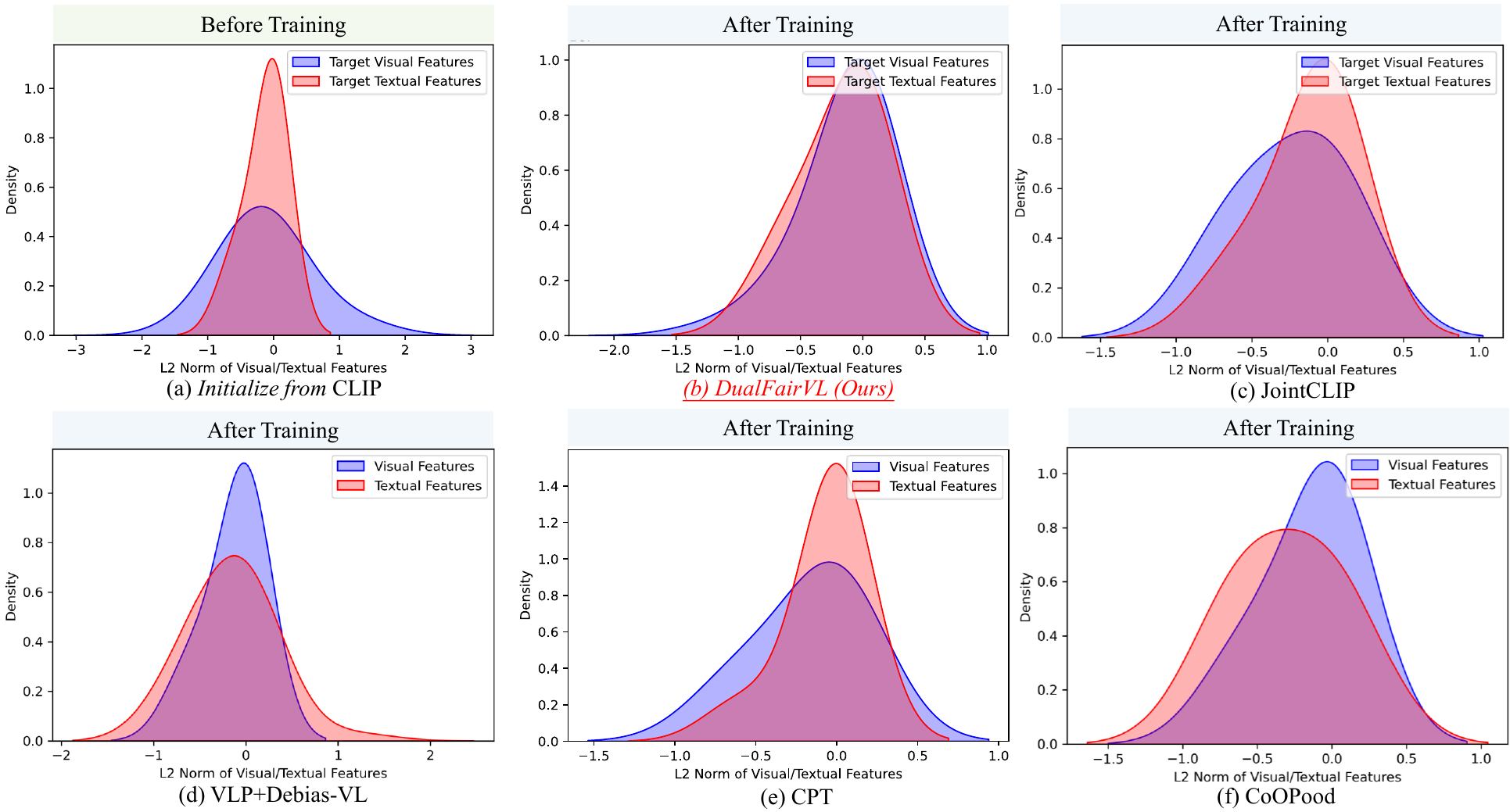}}
\caption{Kernel density estimation of feature distributions, showing the visual-text feature distributions (a) before training and after training with (b) DualFairVL, (c) JointCLIP~\cite{jointclip}, (d) VLP+Debias-VL~\cite{debiasvl}, (e) CPT~\cite{cpt} and (f) CoOPood~\cite{coopood}.}
\label{kde}
\end{figure*}

\subsection{Analysis of Prompt Design Choices}

\subsubsection{Prompt Length} 
Table~\ref{tab:prompt_analysis} (top) shows that increasing prompt length from 2 to 4 improves both AUC and fairness. However, further extending the length to 8 or 16 introduces redundancy, leading to overfitting and weakened fairness regulation. A length of 4 achieves the most balanced trade-off between generalization and fairness.

\subsubsection{Prompt Sharing Strategies} 
Table~\ref{tab:prompt_analysis} (bottom) evaluates different prompt-sharing strategies within the hypernetwork $H(\cdot)$. Layer-aware sharing captures general patterns but lacks instance-level specificity, whereas instance-aware sharing effectively encodes subject-specific cues but neglects broader contextual information. Our combined sharing strategy integrates both, enabling bidirectional feature flow and stronger cross-layer alignment. This results in consistently superior fairness and accuracy across domains.

\subsubsection{Prompt Types and Depth}
Fig.~\ref{prompt_depth} analyzes prompt type and embedding depth when using $H(\cdot)$ for disentanglement. Layer-aware prompts (red) benefit from deeper embedding, capturing hierarchical structures but lacking task-specific guidance. Instance-aware prompts (green) are most effective at the final layer, where task semantics dominate, but can hinder learning when injected too early. DualFairVL (blue) adopts a hybrid design: shallow layers (1st–11th) use layer-aware prompts, while the final layer employs instance-aware prompts. This design ensures that each layer integrates both general and task-specific signals, enhancing generalization and mitigating cross-modal bias.

\subsection{Robustness to Training Data Ratios}
We further test DualFairVL under varying training data ratios (1\%, 5\%, 10\%, 20\%, 50\%, and 100\%) on CheXpert→MIMIC-CXR, keeping the test set fixed. As shown in Fig.~\ref{training_ratio}, the model retains reasonable AUC with as little as 1\% of the training data, although fairness is initially limited. Performance improves steadily with more data, and with only 5\% training data the AUC is within 3\% of the full-data setting while fairness remains stable. These results highlight the data efficiency of our framework, underscoring its suitability for resource-constrained medical applications.

\subsection{Kernel Density Estimation}
Fig.~\ref{kde} visualizes cross-modal feature alignment via kernel density estimation. The pretrained CLIP model (a) shows substantial misalignment, reflecting inherent bias. Text-only debiasing (VLP+Debias-VL\cite{debiasvl}, d) and vision-only debiasing (CoOPood~\cite{coopood}, f) partially reduce bias but fail to achieve full alignment. CPT~\cite{cpt} (e) improves alignment but leaves residual sensitive leakage due to the absence of explicit disentanglement. JointCLIP~\cite{jointclip} (c) applies symmetric linear debiasing but struggles with acquisition-related shifts and multi-scale entanglement. In contrast, DualFairVL (b) achieves substantially improved alignment between debiased visual and textual features, validating the effectiveness of dual-modal prompt-based debiasing.

\begin{figure*}[h]
\centerline{\includegraphics[width=1.0\linewidth]{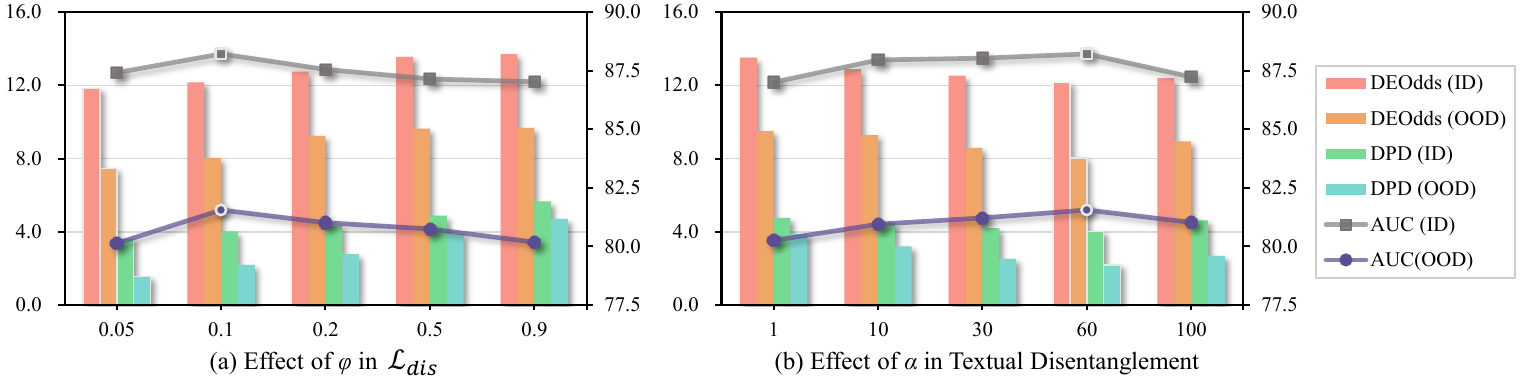}}
\caption{(a) Effect of $\varphi$ in $\mathcal{L}_{dis}$ on the trade-off between fairness and accuracy.
(b) Effect of $\alpha$ in the textual debiasing module on preserving semantic information while removing sensitive attributes.}
\label{hyperparameter}
\end{figure*}

\subsection{Selection of $\varphi$ in Visual Debiasing}
The left panel of Table~\ref{hyperparameter} examines the influence of the temperature coefficient $\varphi$ on $\mathcal{L}_{com}$ and $\mathcal{L}_{sep}$. When $\varphi=0.05$, prototypes cluster tightly, improving fairness but reducing AUC due to limited feature flexibility. For $\varphi \geq 0.2$, both fairness and AUC degrade, reflecting weakened disentanglement. The best trade-off occurs at $\varphi=0.1$, which balances compactness and flexibility, yielding the highest AUC with minimal fairness disparity.

\subsection{Effect of $\alpha$ in Textual Debiasing}
The right panel of Table~\ref{hyperparameter} analyzes the ridge coefficient $\alpha$ in the text-side debiasing projector $Proj(\cdot)$. Optimal fairness (lowest DEOdds and DPD) occurs at $\alpha=60$ in both in-distribution and OOD settings. Excessively large $\alpha$ overemphasizes semantic preservation, leaving residual sensitive components and reducing fairness. Conversely, overly small $\alpha$ aggressively removes sensitive signals but also erodes task-relevant semantics, harming both fairness and accuracy. Performance remains relatively stable around the optimum, suggesting robustness to $\alpha$ within a broad range.

\section{Conclusion}  
In this work, we propose DualFairVL, a text-guided dual-modal debiasing framework for fairness-aware medical image analysis. By constructing orthogonal text anchors, employing cross-modal attention, and disentangling sensitive and target attributes through a hypernetwork with prototype-based regularization, our method achieves robust dual-modal alignment while mitigating bias. Extensive experiments across five in-distribution and three out-of-distribution benchmarks demonstrate that DualFairVL consistently improves both fairness and accuracy, outperforming full-parameter fine-tuning and parameter-efficient baselines with only 3.6M trainable parameters. Beyond setting a new state of the art in fairness-aware domain generalization, our framework is data-efficient and scalable, making it well suited for real-world clinical deployment.

\textbf{Limitations and Future Work.} Our current framework builds on the fine-tuning paradigm of CLIP and has not yet been extended to other vision–language foundation models. Future directions include developing a more general dual-modal debiasing strategy applicable across different VLMs, enabling broader clinical applicability. We also aim to explore more complex and composite sensitive attributes, address scenarios with limited or incomplete annotations, and further improve efficiency and scalability for real-time clinical deployment.

\bmhead{Acknowledgements}

This work was supported by the National Natural Science Foundation
of China under Grant 92470101.

\section*{Declarations}

\begin{itemize}

\item Data availability: The data used in this study are all available at~\cite{fitzpatrick17k,kovalyk2022papila,harvardglaucoma,ham10000,chexpert,bcn,msk,mimiccxr}. 
\end{itemize}




\bibliography{ijcv}


\begin{thebibliography}{59}
\ifx \bisbn   \undefined \def \bisbn  #1{ISBN #1}\fi
\ifx \binits  \undefined \def \binits#1{#1}\fi
\ifx \bauthor  \undefined \def \bauthor#1{#1}\fi
\ifx \batitle  \undefined \def \batitle#1{#1}\fi
\ifx \bjtitle  \undefined \def \bjtitle#1{#1}\fi
\ifx \bvolume  \undefined \def \bvolume#1{\textbf{#1}}\fi
\ifx \byear  \undefined \def \byear#1{#1}\fi
\ifx \bissue  \undefined \def \bissue#1{#1}\fi
\ifx \bfpage  \undefined \def \bfpage#1{#1}\fi
\ifx \blpage  \undefined \def \blpage #1{#1}\fi
\ifx \burl  \undefined \def \burl#1{\textsf{#1}}\fi
\ifx \doiurl  \undefined \def \doiurl#1{\url{https://doi.org/#1}}\fi
\ifx \betal  \undefined \def \betal{\textit{et al.}}\fi
\ifx \binstitute  \undefined \def \binstitute#1{#1}\fi
\ifx \binstitutionaled  \undefined \def \binstitutionaled#1{#1}\fi
\ifx \bctitle  \undefined \def \bctitle#1{#1}\fi
\ifx \beditor  \undefined \def \beditor#1{#1}\fi
\ifx \bpublisher  \undefined \def \bpublisher#1{#1}\fi
\ifx \bbtitle  \undefined \def \bbtitle#1{#1}\fi
\ifx \bedition  \undefined \def \bedition#1{#1}\fi
\ifx \bseriesno  \undefined \def \bseriesno#1{#1}\fi
\ifx \blocation  \undefined \def \blocation#1{#1}\fi
\ifx \bsertitle  \undefined \def \bsertitle#1{#1}\fi
\ifx \bsnm \undefined \def \bsnm#1{#1}\fi
\ifx \bsuffix \undefined \def \bsuffix#1{#1}\fi
\ifx \bparticle \undefined \def \bparticle#1{#1}\fi
\ifx \barticle \undefined \def \barticle#1{#1}\fi
\bibcommenthead
\ifx \bconfdate \undefined \def \bconfdate #1{#1}\fi
\ifx \botherref \undefined \def \botherref #1{#1}\fi
\ifx \url \undefined \def \url#1{\textsf{#1}}\fi
\ifx \bchapter \undefined \def \bchapter#1{#1}\fi
\ifx \bbook \undefined \def \bbook#1{#1}\fi
\ifx \bcomment \undefined \def \bcomment#1{#1}\fi
\ifx \oauthor \undefined \def \oauthor#1{#1}\fi
\ifx \citeauthoryear \undefined \def \citeauthoryear#1{#1}\fi
\ifx \endbibitem  \undefined \def \endbibitem {}\fi
\ifx \bconflocation  \undefined \def \bconflocation#1{#1}\fi
\ifx \arxivurl  \undefined \def \arxivurl#1{\textsf{#1}}\fi
\csname PreBibitemsHook\endcsname

\bibitem[\protect\citeauthoryear{Zong et~al.}{2023}]{zong2022medfair}
\begin{bchapter}
\bauthor{\bsnm{Zong}, \binits{Y.}},
\bauthor{\bsnm{Yang}, \binits{Y.}},
\bauthor{\bsnm{Hospedales}, \binits{T.}}:
\bctitle{Medfair: benchmarking fairness for medical imaging}.
In: \bbtitle{ICLR}
(\byear{2023})
\end{bchapter}
\endbibitem

\bibitem[\protect\citeauthoryear{Wang et~al.}{2023}]{robustfairness}
\begin{botherref}
\oauthor{\bsnm{Wang}, \binits{H.}},
\oauthor{\bsnm{Hong}, \binits{J.}},
\oauthor{\bsnm{Zhou}, \binits{J.}},
\oauthor{\bsnm{Wang}, \binits{Z.}}:
How robust is your fairness? evaluating and sustaining fairness under unseen distribution shifts.
Transactions on machine learning research
(2023)
\end{botherref}
\endbibitem

\bibitem[\protect\citeauthoryear{Dosovitskiy et~al.}{2021}]{vit}
\begin{bchapter}
\bauthor{\bsnm{Dosovitskiy}, \binits{A.}},
\bauthor{\bsnm{Beyer}, \binits{L.}},
\bauthor{\bsnm{Kolesnikov}, \binits{A.}},
\bauthor{\bsnm{Weissenborn}, \binits{D.}},
\bauthor{\bsnm{Zhai}, \binits{X.}},
\bauthor{\bsnm{Unterthiner}, \binits{T.}},
\bauthor{\bsnm{Dehghani}, \binits{M.}},
\bauthor{\bsnm{Minderer}, \binits{M.}},
\bauthor{\bsnm{Heigold}, \binits{G.}},
\bauthor{\bsnm{Gelly}, \binits{S.}}, \betal:
\bctitle{An image is worth 16x16 words: Transformers for image recognition at scale}.
In: \bbtitle{ICLR}
(\byear{2021})
\end{bchapter}
\endbibitem

\bibitem[\protect\citeauthoryear{Radford et~al.}{2021}]{clip}
\begin{bchapter}
\bauthor{\bsnm{Radford}, \binits{A.}},
\bauthor{\bsnm{Kim}, \binits{J.W.}},
\bauthor{\bsnm{Hallacy}, \binits{C.}},
\bauthor{\bsnm{Ramesh}, \binits{A.}},
\bauthor{\bsnm{Goh}, \binits{G.}},
\bauthor{\bsnm{Agarwal}, \binits{S.}},
\bauthor{\bsnm{Sastry}, \binits{G.}},
\bauthor{\bsnm{Askell}, \binits{A.}},
\bauthor{\bsnm{Mishkin}, \binits{P.}},
\bauthor{\bsnm{Clark}, \binits{J.}}, \betal:
\bctitle{Learning transferable visual models from natural language supervision}.
In: \bbtitle{ICML},
pp. \bfpage{8748}--\blpage{8763}
(\byear{2021})
\end{bchapter}
\endbibitem

\bibitem[\protect\citeauthoryear{Jin et~al.}{2024}]{fairmedfm}
\begin{barticle}
\bauthor{\bsnm{Jin}, \binits{R.}},
\bauthor{\bsnm{Xu}, \binits{Z.}},
\bauthor{\bsnm{Zhong}, \binits{Y.}},
\bauthor{\bsnm{Yao}, \binits{Q.}},
\bauthor{\bsnm{QI}, \binits{D.}},
\bauthor{\bsnm{Zhou}, \binits{S.K.}},
\bauthor{\bsnm{Li}, \binits{X.}}:
\batitle{Fairmedfm: fairness benchmarking for medical imaging foundation models}.
\bjtitle{NeurIPS}
\bvolume{37},
\bfpage{111318}--\blpage{111357}
(\byear{2024})
\end{barticle}
\endbibitem

\bibitem[\protect\citeauthoryear{Park and Byun}{2024}]{fairvpt}
\begin{bchapter}
\bauthor{\bsnm{Park}, \binits{S.}},
\bauthor{\bsnm{Byun}, \binits{H.}}:
\bctitle{Fair-vpt: Fair visual prompt tuning for image classification}.
In: \bbtitle{CVPR},
pp. \bfpage{12268}--\blpage{12278}
(\byear{2024})
\end{bchapter}
\endbibitem

\bibitem[\protect\citeauthoryear{Tian et~al.}{2024}]{fairdomain}
\begin{bchapter}
\bauthor{\bsnm{Tian}, \binits{Y.}},
\bauthor{\bsnm{Wen}, \binits{C.}},
\bauthor{\bsnm{Shi}, \binits{M.}},
\bauthor{\bsnm{Afzal}, \binits{M.M.}},
\bauthor{\bsnm{Huang}, \binits{H.}},
\bauthor{\bsnm{Khan}, \binits{M.O.}},
\bauthor{\bsnm{Luo}, \binits{Y.}},
\bauthor{\bsnm{Fang}, \binits{Y.}},
\bauthor{\bsnm{Wang}, \binits{M.}}:
\bctitle{Fairdomain: Achieving fairness in cross-domain medical image segmentation and classification}.
In: \bbtitle{ECCV},
pp. \bfpage{251}--\blpage{271}
(\byear{2024})
\end{bchapter}
\endbibitem

\bibitem[\protect\citeauthoryear{Dutt et~al.}{2024}]{fairtune}
\begin{bchapter}
\bauthor{\bsnm{Dutt}, \binits{R.}},
\bauthor{\bsnm{Bohdal}, \binits{O.}},
\bauthor{\bsnm{Tsaftaris}, \binits{S.A.}},
\bauthor{\bsnm{Hospedales}, \binits{T.}}:
\bctitle{Fairtune: Optimizing parameter efficient fine tuning for fairness in medical image analysis}.
In: \bbtitle{ICLR}
(\byear{2024})
\end{bchapter}
\endbibitem

\bibitem[\protect\citeauthoryear{Chuang et~al.}{2023}]{debiasvl}
\begin{botherref}
\oauthor{\bsnm{Chuang}, \binits{C.-Y.}},
\oauthor{\bsnm{Jampani}, \binits{V.}},
\oauthor{\bsnm{Li}, \binits{Y.}},
\oauthor{\bsnm{Torralba}, \binits{A.}},
\oauthor{\bsnm{Jegelka}, \binits{S.}}:
Debiasing vision-language models via biased prompts.
arXiv preprint arXiv:2302.00070
(2023)
\end{botherref}
\endbibitem

\bibitem[\protect\citeauthoryear{Liang et~al.}{2020}]{biasdirection}
\begin{bchapter}
\bauthor{\bsnm{Liang}, \binits{P.P.}},
\bauthor{\bsnm{Li}, \binits{I.M.}},
\bauthor{\bsnm{Zheng}, \binits{E.}},
\bauthor{\bsnm{Lim}, \binits{Y.C.}},
\bauthor{\bsnm{Salakhutdinov}, \binits{R.}},
\bauthor{\bsnm{Morency}, \binits{L.}}:
\bctitle{Towards debiasing sentence representations}.
In: \bbtitle{ACL},
pp. \bfpage{5502}--\blpage{5515}
(\byear{2020})
\end{bchapter}
\endbibitem

\bibitem[\protect\citeauthoryear{Deng et~al.}{2023}]{orthfairness}
\begin{bchapter}
\bauthor{\bsnm{Deng}, \binits{W.}},
\bauthor{\bsnm{Zhong}, \binits{Y.}},
\bauthor{\bsnm{Dou}, \binits{Q.}},
\bauthor{\bsnm{Li}, \binits{X.}}:
\bctitle{On fairness of medical image classification with multiple sensitive attributes via learning orthogonal representations}.
In: \bbtitle{IPMI},
pp. \bfpage{158}--\blpage{169}
(\byear{2023})
\end{bchapter}
\endbibitem

\bibitem[\protect\citeauthoryear{Molahasani et~al.}{2025}]{prism}
\begin{bchapter}
\bauthor{\bsnm{Molahasani}, \binits{M.}},
\bauthor{\bsnm{Motamedi}, \binits{A.}},
\bauthor{\bsnm{Greenspan}, \binits{M.}},
\bauthor{\bsnm{Kim}, \binits{I.-M.}},
\bauthor{\bsnm{Etemad}, \binits{A.}}:
\bctitle{Prism: Reducing spurious implicit biases in vision-language models with llm-guided embedding projection}.
In: \bbtitle{ICCV}
(\byear{2025})
\end{bchapter}
\endbibitem

\bibitem[\protect\citeauthoryear{Luo et~al.}{2024}]{fairclip}
\begin{bchapter}
\bauthor{\bsnm{Luo}, \binits{Y.}},
\bauthor{\bsnm{Shi}, \binits{M.}},
\bauthor{\bsnm{Khan}, \binits{M.O.}},
\bauthor{\bsnm{Afzal}, \binits{M.M.}},
\bauthor{\bsnm{Huang}, \binits{H.}},
\bauthor{\bsnm{Yuan}, \binits{S.}},
\bauthor{\bsnm{Tian}, \binits{Y.}},
\bauthor{\bsnm{Song}, \binits{L.}},
\bauthor{\bsnm{Kouhana}, \binits{A.}},
\bauthor{\bsnm{Elze}, \binits{T.}}, \betal:
\bctitle{Fairclip: Harnessing fairness in vision-language learning}.
In: \bbtitle{CVPR},
pp. \bfpage{12289}--\blpage{12301}
(\byear{2024})
\end{bchapter}
\endbibitem

\bibitem[\protect\citeauthoryear{Berg et~al.}{2022}]{itc}
\begin{bchapter}
\bauthor{\bsnm{Berg}, \binits{H.}},
\bauthor{\bsnm{Hall}, \binits{S.M.}},
\bauthor{\bsnm{Bhalgat}, \binits{Y.}},
\bauthor{\bsnm{Yang}, \binits{W.}},
\bauthor{\bsnm{Kirk}, \binits{H.R.}},
\bauthor{\bsnm{Shtedritski}, \binits{A.}},
\bauthor{\bsnm{Bain}, \binits{M.}}:
\bctitle{A prompt array keeps the bias away: Debiasing vision-language models with adversarial learning}.
In: \bbtitle{AACL-IJCNLP},
pp. \bfpage{806}--\blpage{822}
(\byear{2022})
\end{bchapter}
\endbibitem

\bibitem[\protect\citeauthoryear{Seth et~al.}{2023}]{dear}
\begin{bchapter}
\bauthor{\bsnm{Seth}, \binits{A.}},
\bauthor{\bsnm{Hemani}, \binits{M.}},
\bauthor{\bsnm{Agarwal}, \binits{C.}}:
\bctitle{Dear: Debiasing vision-language models with additive residuals}.
In: \bbtitle{CVPR},
pp. \bfpage{6820}--\blpage{6829}
(\byear{2023})
\end{bchapter}
\endbibitem

\bibitem[\protect\citeauthoryear{Zhang and R{\'e}}{2022}]{ca}
\begin{barticle}
\bauthor{\bsnm{Zhang}, \binits{M.}},
\bauthor{\bsnm{R{\'e}}, \binits{C.}}:
\batitle{Contrastive adapters for foundation model group robustness}.
\bjtitle{NeurIPS}
\bvolume{35},
\bfpage{21682}--\blpage{21697}
(\byear{2022})
\end{barticle}
\endbibitem

\bibitem[\protect\citeauthoryear{Zhang et~al.}{2024}]{coopood}
\begin{bchapter}
\bauthor{\bsnm{Zhang}, \binits{J.}},
\bauthor{\bsnm{Ma}, \binits{X.}},
\bauthor{\bsnm{Guo}, \binits{S.}},
\bauthor{\bsnm{Li}, \binits{P.}},
\bauthor{\bsnm{Xu}, \binits{W.}},
\bauthor{\bsnm{Tang}, \binits{X.}},
\bauthor{\bsnm{Hong}, \binits{Z.}}:
\bctitle{Amend to alignment: decoupled prompt tuning for mitigating spurious correlation in vision-language models}.
In: \bbtitle{ICML}
(\byear{2024})
\end{bchapter}
\endbibitem

\bibitem[\protect\citeauthoryear{You et~al.}{2024}]{cfr}
\begin{bchapter}
\bauthor{\bsnm{You}, \binits{C.}},
\bauthor{\bsnm{Mint}, \binits{Y.}},
\bauthor{\bsnm{Dai}, \binits{W.}},
\bauthor{\bsnm{Sekhon}, \binits{J.S.}},
\bauthor{\bsnm{Staib}, \binits{L.}},
\bauthor{\bsnm{Duncan}, \binits{J.S.}}:
\bctitle{Calibrating multi-modal representations: A pursuit of group robustness without annotations}.
In: \bbtitle{CVPR},
pp. \bfpage{26140}--\blpage{26150}
(\byear{2024})
\end{bchapter}
\endbibitem

\bibitem[\protect\citeauthoryear{Yang et~al.}{2023}]{mitigating_sc}
\begin{bchapter}
\bauthor{\bsnm{Yang}, \binits{Y.}},
\bauthor{\bsnm{Nushi}, \binits{B.}},
\bauthor{\bsnm{Palangi}, \binits{H.}},
\bauthor{\bsnm{Mirzasoleiman}, \binits{B.}}:
\bctitle{Mitigating spurious correlations in multi-modal models during fine-tuning}.
In: \bbtitle{ICML},
pp. \bfpage{39365}--\blpage{39379}
(\byear{2023})
\end{bchapter}
\endbibitem

\bibitem[\protect\citeauthoryear{Jin et~al.}{2022}]{goodprompt}
\begin{bchapter}
\bauthor{\bsnm{Jin}, \binits{W.}},
\bauthor{\bsnm{Cheng}, \binits{Y.}},
\bauthor{\bsnm{Shen}, \binits{Y.}},
\bauthor{\bsnm{Chen}, \binits{W.}},
\bauthor{\bsnm{Ren}, \binits{X.}}:
\bctitle{A good prompt is worth millions of parameters: Low-resource prompt-based learning for vision-language models}.
In: \bbtitle{AMACL},
pp. \bfpage{2763}--\blpage{2775}
(\byear{2022})
\end{bchapter}
\endbibitem

\bibitem[\protect\citeauthoryear{Wasim et~al.}{2023}]{vita_clip}
\begin{bchapter}
\bauthor{\bsnm{Wasim}, \binits{S.T.}},
\bauthor{\bsnm{Naseer}, \binits{M.}},
\bauthor{\bsnm{Khan}, \binits{S.}},
\bauthor{\bsnm{Khan}, \binits{F.S.}},
\bauthor{\bsnm{Shah}, \binits{M.}}:
\bctitle{Vita-clip: Video and text adaptive clip via multimodal prompting}.
In: \bbtitle{CVPR},
pp. \bfpage{23034}--\blpage{23044}
(\byear{2023})
\end{bchapter}
\endbibitem

\bibitem[\protect\citeauthoryear{Huang et~al.}{2023}]{vop}
\begin{bchapter}
\bauthor{\bsnm{Huang}, \binits{S.}},
\bauthor{\bsnm{Gong}, \binits{B.}},
\bauthor{\bsnm{Pan}, \binits{Y.}},
\bauthor{\bsnm{Jiang}, \binits{J.}},
\bauthor{\bsnm{Lv}, \binits{Y.}},
\bauthor{\bsnm{Li}, \binits{Y.}},
\bauthor{\bsnm{Wang}, \binits{D.}}:
\bctitle{Vop: Text-video co-operative prompt tuning for cross-modal retrieval}.
In: \bbtitle{CVPR},
pp. \bfpage{6565}--\blpage{6574}
(\byear{2023})
\end{bchapter}
\endbibitem

\bibitem[\protect\citeauthoryear{Khattak et~al.}{2023}]{maple}
\begin{bchapter}
\bauthor{\bsnm{Khattak}, \binits{M.U.}},
\bauthor{\bsnm{Rasheed}, \binits{H.}},
\bauthor{\bsnm{Maaz}, \binits{M.}},
\bauthor{\bsnm{Khan}, \binits{S.}},
\bauthor{\bsnm{Khan}, \binits{F.S.}}:
\bctitle{Maple: Multi-modal prompt learning}.
In: \bbtitle{CVPR},
pp. \bfpage{19113}--\blpage{19122}
(\byear{2023})
\end{bchapter}
\endbibitem

\bibitem[\protect\citeauthoryear{Irvin et~al.}{2019}]{chexpert}
\begin{bchapter}
\bauthor{\bsnm{Irvin}, \binits{J.}},
\bauthor{\bsnm{Rajpurkar}, \binits{P.}},
\bauthor{\bsnm{Ko}, \binits{M.}},
\bauthor{\bsnm{Yu}, \binits{Y.}},
\bauthor{\bsnm{Ciurea-Ilcus}, \binits{S.}},
\bauthor{\bsnm{Chute}, \binits{C.}},
\bauthor{\bsnm{Marklund}, \binits{H.}},
\bauthor{\bsnm{Haghgoo}, \binits{B.}},
\bauthor{\bsnm{Ball}, \binits{R.}},
\bauthor{\bsnm{Shpanskaya}, \binits{K.}}, \betal:
\bctitle{Chexpert: A large chest radiograph dataset with uncertainty labels and expert comparison}.
In: \bbtitle{AAAI},
vol. \bseriesno{33},
pp. \bfpage{590}--\blpage{597}
(\byear{2019})
\end{bchapter}
\endbibitem

\bibitem[\protect\citeauthoryear{Johnson et~al.}{2019}]{mimiccxr}
\begin{barticle}
\bauthor{\bsnm{Johnson}, \binits{A.E.}},
\bauthor{\bsnm{Pollard}, \binits{T.J.}},
\bauthor{\bsnm{Berkowitz}, \binits{S.J.}},
\bauthor{\bsnm{Greenbaum}, \binits{N.R.}},
\bauthor{\bsnm{Lungren}, \binits{M.P.}},
\bauthor{\bsnm{Deng}, \binits{C.-y.}},
\bauthor{\bsnm{Mark}, \binits{R.G.}},
\bauthor{\bsnm{Horng}, \binits{S.}}:
\batitle{Mimic-cxr, a de-identified publicly available database of chest radiographs with free-text reports}.
\bjtitle{Scientific data}
\bvolume{6}(\bissue{1}),
\bfpage{317}
(\byear{2019})
\end{barticle}
\endbibitem

\bibitem[\protect\citeauthoryear{Groh et~al.}{2021}]{fitzpatrick17k}
\begin{bchapter}
\bauthor{\bsnm{Groh}, \binits{M.}},
\bauthor{\bsnm{Harris}, \binits{C.}},
\bauthor{\bsnm{Soenksen}, \binits{L.}},
\bauthor{\bsnm{Lau}, \binits{F.}},
\bauthor{\bsnm{Han}, \binits{R.}},
\bauthor{\bsnm{Kim}, \binits{A.}},
\bauthor{\bsnm{Koochek}, \binits{A.}},
\bauthor{\bsnm{Badri}, \binits{O.}}:
\bctitle{Evaluating deep neural networks trained on clinical images in dermatology with the fitzpatrick 17k dataset}.
In: \bbtitle{CVPR},
pp. \bfpage{1820}--\blpage{1828}
(\byear{2021})
\end{bchapter}
\endbibitem

\bibitem[\protect\citeauthoryear{Kovalyk et~al.}{2022}]{kovalyk2022papila}
\begin{barticle}
\bauthor{\bsnm{Kovalyk}, \binits{O.}},
\bauthor{\bsnm{Morales-S{\'a}nchez}, \binits{J.}},
\bauthor{\bsnm{Verd{\'u}-Monedero}, \binits{R.}},
\bauthor{\bsnm{Sell{\'e}s-Navarro}, \binits{I.}},
\bauthor{\bsnm{Palaz{\'o}n-Cabanes}, \binits{A.}},
\bauthor{\bsnm{Sancho-G{\'o}mez}, \binits{J.-L.}}:
\batitle{Papila: Dataset with fundus images and clinical data of both eyes of the same patient for glaucoma assessment}.
\bjtitle{Scientific Data}
\bvolume{9}(\bissue{1}),
\bfpage{291}
(\byear{2022})
\end{barticle}
\endbibitem

\bibitem[\protect\citeauthoryear{Luo et~al.}{2024}]{harvardglaucoma}
\begin{botherref}
\oauthor{\bsnm{Luo}, \binits{Y.}},
\oauthor{\bsnm{Tian}, \binits{Y.}},
\oauthor{\bsnm{Shi}, \binits{M.}},
\oauthor{\bsnm{Pasquale}, \binits{L.R.}},
\oauthor{\bsnm{Shen}, \binits{L.Q.}},
\oauthor{\bsnm{Zebardast}, \binits{N.}},
\oauthor{\bsnm{Elze}, \binits{T.}},
\oauthor{\bsnm{Wang}, \binits{M.}}:
Harvard glaucoma fairness: a retinal nerve disease dataset for fairness learning and fair identity normalization.
IEEE Transactions on Medical Imaging
(2024)
\end{botherref}
\endbibitem

\bibitem[\protect\citeauthoryear{Tschandl et~al.}{2018}]{ham10000}
\begin{barticle}
\bauthor{\bsnm{Tschandl}, \binits{P.}},
\bauthor{\bsnm{Rosendahl}, \binits{C.}},
\bauthor{\bsnm{Kittler}, \binits{H.}}:
\batitle{The ham10000 dataset, a large collection of multi-source dermatoscopic images of common pigmented skin lesions}.
\bjtitle{Scientific data}
\bvolume{5}(\bissue{1}),
\bfpage{1}--\blpage{9}
(\byear{2018})
\end{barticle}
\endbibitem

\bibitem[\protect\citeauthoryear{Puyol-Ant{\'o}n et~al.}{2021}]{pre_mrfairness}
\begin{bchapter}
\bauthor{\bsnm{Puyol-Ant{\'o}n}, \binits{E.}},
\bauthor{\bsnm{Ruijsink}, \binits{B.}},
\bauthor{\bsnm{Piechnik}, \binits{S.K.}},
\bauthor{\bsnm{Neubauer}, \binits{S.}},
\bauthor{\bsnm{Petersen}, \binits{S.E.}},
\bauthor{\bsnm{Razavi}, \binits{R.}},
\bauthor{\bsnm{King}, \binits{A.P.}}:
\bctitle{Fairness in cardiac mr image analysis: an investigation of bias due to data imbalance in deep learning based segmentation}.
In: \bbtitle{MICCAI},
pp. \bfpage{413}--\blpage{423}
(\byear{2021})
\end{bchapter}
\endbibitem

\bibitem[\protect\citeauthoryear{Park et~al.}{2022}]{fscl}
\begin{bchapter}
\bauthor{\bsnm{Park}, \binits{S.}},
\bauthor{\bsnm{Lee}, \binits{J.}},
\bauthor{\bsnm{Lee}, \binits{P.}},
\bauthor{\bsnm{Hwang}, \binits{S.}},
\bauthor{\bsnm{Kim}, \binits{D.}},
\bauthor{\bsnm{Byun}, \binits{H.}}:
\bctitle{Fair contrastive learning for facial attribute classification}.
In: \bbtitle{CVPR},
pp. \bfpage{10389}--\blpage{10398}
(\byear{2022})
\end{bchapter}
\endbibitem

\bibitem[\protect\citeauthoryear{Sagawa et~al.}{2020}]{groupdro}
\begin{bchapter}
\bauthor{\bsnm{Sagawa}, \binits{S.}},
\bauthor{\bsnm{Koh}, \binits{P.W.}},
\bauthor{\bsnm{Hashimoto}, \binits{T.B.}},
\bauthor{\bsnm{Liang}, \binits{P.}}:
\bctitle{Distributionally robust neural networks for group shifts: On the importance of regularization for worst-case generalization}.
In: \bbtitle{ICLR},
pp. \bfpage{10389}--\blpage{10398}
(\byear{2020})
\end{bchapter}
\endbibitem

\bibitem[\protect\citeauthoryear{Raff and Sylvester}{2018}]{at_vit}
\begin{bchapter}
\bauthor{\bsnm{Raff}, \binits{E.}},
\bauthor{\bsnm{Sylvester}, \binits{J.}}:
\bctitle{Gradient reversal against discrimination: A fair neural network learning approach}.
In: \bbtitle{DSAA},
pp. \bfpage{189}--\blpage{198}
(\byear{2018})
\end{bchapter}
\endbibitem

\bibitem[\protect\citeauthoryear{Marcinkevics et~al.}{2022}]{postdebiasing}
\begin{bchapter}
\bauthor{\bsnm{Marcinkevics}, \binits{R.}},
\bauthor{\bsnm{Ozkan}, \binits{E.}},
\bauthor{\bsnm{Vogt}, \binits{J.E.}}:
\bctitle{Debiasing deep chest x-ray classifiers using intra-and post-processing methods}.
In: \bbtitle{Machine Learning for Healthcare Conference},
pp. \bfpage{504}--\blpage{536}
(\byear{2022})
\end{bchapter}
\endbibitem

\bibitem[\protect\citeauthoryear{Wu et~al.}{2022}]{fairprune}
\begin{bchapter}
\bauthor{\bsnm{Wu}, \binits{Y.}},
\bauthor{\bsnm{Zeng}, \binits{D.}},
\bauthor{\bsnm{Xu}, \binits{X.}},
\bauthor{\bsnm{Shi}, \binits{Y.}},
\bauthor{\bsnm{Hu}, \binits{J.}}:
\bctitle{Fairprune: Achieving fairness through pruning for dermatological disease diagnosis}.
In: \bbtitle{MICCAI},
pp. \bfpage{743}--\blpage{753}
(\byear{2022})
\end{bchapter}
\endbibitem

\bibitem[\protect\citeauthoryear{Zhu et~al.}{2019}]{dgfair_aligning}
\begin{bchapter}
\bauthor{\bsnm{Zhu}, \binits{Y.}},
\bauthor{\bsnm{Zhuang}, \binits{F.}},
\bauthor{\bsnm{Wang}, \binits{D.}}:
\bctitle{Aligning domain-specific distribution and classifier for cross-domain classification from multiple sources}.
In: \bbtitle{AAAI},
vol. \bseriesno{33},
pp. \bfpage{5989}--\blpage{5996}
(\byear{2019})
\end{bchapter}
\endbibitem

\bibitem[\protect\citeauthoryear{Truong et~al.}{2023}]{dgfair_fredom}
\begin{bchapter}
\bauthor{\bsnm{Truong}, \binits{T.-D.}},
\bauthor{\bsnm{Le}, \binits{N.}},
\bauthor{\bsnm{Raj}, \binits{B.}},
\bauthor{\bsnm{Cothren}, \binits{J.}},
\bauthor{\bsnm{Luu}, \binits{K.}}:
\bctitle{Fredom: Fairness domain adaptation approach to semantic scene understanding}.
In: \bbtitle{CVPR},
pp. \bfpage{19988}--\blpage{19997}
(\byear{2023})
\end{bchapter}
\endbibitem

\bibitem[\protect\citeauthoryear{Xiao et~al.}{2023}]{medmae}
\begin{bchapter}
\bauthor{\bsnm{Xiao}, \binits{J.}},
\bauthor{\bsnm{Bai}, \binits{Y.}},
\bauthor{\bsnm{Yuille}, \binits{A.}},
\bauthor{\bsnm{Zhou}, \binits{Z.}}:
\bctitle{Delving into masked autoencoders for multi-label thorax disease classification}.
In: \bbtitle{WACV},
pp. \bfpage{3588}--\blpage{3600}
(\byear{2023})
\end{bchapter}
\endbibitem

\bibitem[\protect\citeauthoryear{Oquab et~al.}{2024}]{dinov2}
\begin{botherref}
\oauthor{\bsnm{Oquab}, \binits{M.}},
\oauthor{\bsnm{Darcet}, \binits{T.}},
\oauthor{\bsnm{Moutakanni}, \binits{T.}},
\oauthor{\bsnm{Vo}, \binits{H.}},
\oauthor{\bsnm{Szafraniec}, \binits{M.}},
\oauthor{\bsnm{Khalidov}, \binits{V.}},
\oauthor{\bsnm{Fernandez}, \binits{P.}},
\oauthor{\bsnm{Haziza}, \binits{D.}},
\oauthor{\bsnm{Massa}, \binits{F.}},
\oauthor{\bsnm{El-Nouby}, \binits{A.}}, et al.:
Dinov2: Learning robust visual features without supervision.
Transactions on Machine Learning Research,
2835--8856
(2024)
\end{botherref}
\endbibitem

\bibitem[\protect\citeauthoryear{MH~Nguyen et~al.}{2023}]{medlvm}
\begin{barticle}
\bauthor{\bsnm{MH~Nguyen}, \binits{D.}},
\bauthor{\bsnm{Nguyen}, \binits{H.}},
\bauthor{\bsnm{Diep}, \binits{N.}},
\bauthor{\bsnm{Pham}, \binits{T.N.}},
\bauthor{\bsnm{Cao}, \binits{T.}},
\bauthor{\bsnm{Nguyen}, \binits{B.}},
\bauthor{\bsnm{Swoboda}, \binits{P.}},
\bauthor{\bsnm{Ho}, \binits{N.}},
\bauthor{\bsnm{Albarqouni}, \binits{S.}},
\bauthor{\bsnm{Xie}, \binits{P.}}, \betal:
\batitle{Lvm-med: Learning large-scale self-supervised vision models for medical imaging via second-order graph matching}.
\bjtitle{NeurIPS}
\bvolume{36},
\bfpage{27922}--\blpage{27950}
(\byear{2023})
\end{barticle}
\endbibitem

\bibitem[\protect\citeauthoryear{Wang et~al.}{2022}]{medclip}
\begin{bchapter}
\bauthor{\bsnm{Wang}, \binits{Z.}},
\bauthor{\bsnm{Wu}, \binits{Z.}},
\bauthor{\bsnm{Agarwal}, \binits{D.}},
\bauthor{\bsnm{Sun}, \binits{J.}}:
\bctitle{Medclip: Contrastive learning from unpaired medical images and text}.
In: \bbtitle{EMNLP},
vol. \bseriesno{2022},
pp. \bfpage{3876}--\blpage{3887}
(\byear{2022})
\end{bchapter}
\endbibitem

\bibitem[\protect\citeauthoryear{Zhang et~al.}{2023}]{biomedclip}
\begin{botherref}
\oauthor{\bsnm{Zhang}, \binits{S.}},
\oauthor{\bsnm{Xu}, \binits{Y.}},
\oauthor{\bsnm{Usuyama}, \binits{N.}},
\oauthor{\bsnm{Xu}, \binits{H.}},
\oauthor{\bsnm{Bagga}, \binits{J.}},
\oauthor{\bsnm{Tinn}, \binits{R.}},
\oauthor{\bsnm{Preston}, \binits{S.}},
\oauthor{\bsnm{Rao}, \binits{R.}},
\oauthor{\bsnm{Wei}, \binits{M.}},
\oauthor{\bsnm{Valluri}, \binits{N.}}, et al.:
Biomedclip: a multimodal biomedical foundation model pretrained from fifteen million scientific image-text pairs.
arXiv preprint arXiv:2303.00915
(2023)
\end{botherref}
\endbibitem

\bibitem[\protect\citeauthoryear{Jia et~al.}{2022}]{vpt}
\begin{bchapter}
\bauthor{\bsnm{Jia}, \binits{M.}},
\bauthor{\bsnm{Tang}, \binits{L.}},
\bauthor{\bsnm{Chen}, \binits{B.-C.}},
\bauthor{\bsnm{Cardie}, \binits{C.}},
\bauthor{\bsnm{Belongie}, \binits{S.}},
\bauthor{\bsnm{Hariharan}, \binits{B.}},
\bauthor{\bsnm{Lim}, \binits{S.-N.}}:
\bctitle{Visual prompt tuning}.
In: \bbtitle{ECCV},
pp. \bfpage{709}--\blpage{727}
(\byear{2022})
\end{bchapter}
\endbibitem

\bibitem[\protect\citeauthoryear{Zhou et~al.}{2022}]{coop}
\begin{barticle}
\bauthor{\bsnm{Zhou}, \binits{K.}},
\bauthor{\bsnm{Yang}, \binits{J.}},
\bauthor{\bsnm{Loy}, \binits{C.C.}},
\bauthor{\bsnm{Liu}, \binits{Z.}}:
\batitle{Learning to prompt for vision-language models}.
\bjtitle{International Journal of Computer Vision}
\bvolume{130}(\bissue{9}),
\bfpage{2337}--\blpage{2348}
(\byear{2022})
\end{barticle}
\endbibitem

\bibitem[\protect\citeauthoryear{{Zhou, Kaiyang} et~al.}{2022}]{cocoop}
\begin{bchapter}
\bauthor{\bsnm{{Zhou, Kaiyang}}},
\bauthor{\bsnm{{Yang, Jingkang}}},
\bauthor{\bsnm{{Loy, Chen Change}}},
\bauthor{\bsnm{{Liu, Ziwei}}}:
\bctitle{Conditional prompt learning for vision-language models}.
In: \bbtitle{CVPR},
pp. \bfpage{16816}--\blpage{16825}
(\byear{2022})
\end{bchapter}
\endbibitem

\bibitem[\protect\citeauthoryear{Wang et~al.}{2024}]{mcpl}
\begin{botherref}
\oauthor{\bsnm{Wang}, \binits{P.}},
\oauthor{\bsnm{Zhang}, \binits{H.}},
\oauthor{\bsnm{Yuan}, \binits{Y.}}:
Mcpl: Multi-modal collaborative prompt learning for medical vision-language model.
IEEE Transactions on Medical Imaging
(2024)
\end{botherref}
\endbibitem

\bibitem[\protect\citeauthoryear{Dehdashtian et~al.}{2024}]{fairerclip}
\begin{bchapter}
\bauthor{\bsnm{Dehdashtian}, \binits{S.}},
\bauthor{\bsnm{Wang}, \binits{L.}},
\bauthor{\bsnm{Boddeti}, \binits{V.N.}}:
\bctitle{Fairerclip: Debiasing clip's zero-shot predictions using functions in rkhss}.
In: \bbtitle{ICLR}
(\byear{2024})
\end{bchapter}
\endbibitem

\bibitem[\protect\citeauthoryear{Phan et~al.}{2024}]{cpt}
\begin{bchapter}
\bauthor{\bsnm{Phan}, \binits{H.}},
\bauthor{\bsnm{Wilson}, \binits{A.G.}},
\bauthor{\bsnm{Lei}, \binits{Q.}}:
\bctitle{Controllable prompt tuning for balancing group distributional robustness}.
In: \bbtitle{ICML}
(\byear{2024})
\end{bchapter}
\endbibitem

\bibitem[\protect\citeauthoryear{Zhang et~al.}{2025}]{jointclip}
\begin{bchapter}
\bauthor{\bsnm{Zhang}, \binits{H.}},
\bauthor{\bsnm{Guo}, \binits{Y.}},
\bauthor{\bsnm{Kankanhalli}, \binits{M.}}:
\bctitle{Joint vision-language social bias removal for clip}.
In: \bbtitle{CVPR},
pp. \bfpage{4246}--\blpage{4255}
(\byear{2025})
\end{bchapter}
\endbibitem

\bibitem[\protect\citeauthoryear{Sch{\"o}lkopf et~al.}{2016}]{hsr}
\begin{barticle}
\bauthor{\bsnm{Sch{\"o}lkopf}, \binits{B.}},
\bauthor{\bsnm{Hogg}, \binits{D.W.}},
\bauthor{\bsnm{Wang}, \binits{D.}},
\bauthor{\bsnm{Foreman-Mackey}, \binits{D.}},
\bauthor{\bsnm{Janzing}, \binits{D.}},
\bauthor{\bsnm{Simon-Gabriel}, \binits{C.-J.}},
\bauthor{\bsnm{Peters}, \binits{J.}}:
\batitle{Modeling confounding by half-sibling regression}.
\bjtitle{PNAS}
\bvolume{113}(\bissue{27}),
\bfpage{7391}--\blpage{7398}
(\byear{2016})
\end{barticle}
\endbibitem

\bibitem[\protect\citeauthoryear{Sarhan et~al.}{2020}]{eccvorthfairness}
\begin{bchapter}
\bauthor{\bsnm{Sarhan}, \binits{M.H.}},
\bauthor{\bsnm{Navab}, \binits{N.}},
\bauthor{\bsnm{Eslami}, \binits{A.}},
\bauthor{\bsnm{Albarqouni}, \binits{S.}}:
\bctitle{Fairness by learning orthogonal disentangled representations}.
In: \bbtitle{ECCV},
pp. \bfpage{746}--\blpage{761}
(\byear{2020}).
\bcomment{Springer}
\end{bchapter}
\endbibitem

\bibitem[\protect\citeauthoryear{Mardia and Jupp}{2000}]{vmf}
\begin{bbook}
\bauthor{\bsnm{Mardia}, \binits{K.V.}},
\bauthor{\bsnm{Jupp}, \binits{P.E.}}:
\bbtitle{Directional Statistics}.
\bpublisher{Wiley Online Library},
\blocation{Hoboken, NJ, USA}
(\byear{2000})
\end{bbook}
\endbibitem

\bibitem[\protect\citeauthoryear{Maron et~al.}{2019}]{bin_derm}
\begin{barticle}
\bauthor{\bsnm{Maron}, \binits{R.C.}},
\bauthor{\bsnm{Weichenthal}, \binits{M.}},
\bauthor{\bsnm{Utikal}, \binits{J.S.}},
\bauthor{\bsnm{Hekler}, \binits{A.}},
\bauthor{\bsnm{Berking}, \binits{C.}},
\bauthor{\bsnm{Hauschild}, \binits{A.}},
\bauthor{\bsnm{Enk}, \binits{A.H.}},
\bauthor{\bsnm{Haferkamp}, \binits{S.}},
\bauthor{\bsnm{Klode}, \binits{J.}},
\bauthor{\bsnm{Schadendorf}, \binits{D.}}, \betal:
\batitle{Systematic outperformance of 112 dermatologists in multiclass skin cancer image classification by convolutional neural networks}.
\bjtitle{European Journal of Cancer}
\bvolume{119},
\bfpage{57}--\blpage{65}
(\byear{2019})
\end{barticle}
\endbibitem

\bibitem[\protect\citeauthoryear{Gichoya et~al.}{2022}]{processchexpert}
\begin{barticle}
\bauthor{\bsnm{Gichoya}, \binits{J.W.}},
\bauthor{\bsnm{Banerjee}, \binits{I.}},
\bauthor{\bsnm{Bhimireddy}, \binits{A.R.}},
\bauthor{\bsnm{Burns}, \binits{J.L.}},
\bauthor{\bsnm{Celi}, \binits{L.A.}},
\bauthor{\bsnm{Chen}, \binits{L.-C.}},
\bauthor{\bsnm{Correa}, \binits{R.}},
\bauthor{\bsnm{Dullerud}, \binits{N.}},
\bauthor{\bsnm{Ghassemi}, \binits{M.}},
\bauthor{\bsnm{Huang}, \binits{S.-C.}}, \betal:
\batitle{Ai recognition of patient race in medical imaging: a modelling study}.
\bjtitle{The Lancet Digital Health}
\bvolume{4}(\bissue{6}),
\bfpage{406}--\blpage{414}
(\byear{2022})
\end{barticle}
\endbibitem

\bibitem[\protect\citeauthoryear{Hern{\'a}ndez-P{\'e}rez et~al.}{2024}]{bcn}
\begin{barticle}
\bauthor{\bsnm{Hern{\'a}ndez-P{\'e}rez}, \binits{C.}},
\bauthor{\bsnm{Combalia}, \binits{M.}},
\bauthor{\bsnm{Podlipnik}, \binits{S.}},
\bauthor{\bsnm{Codella}, \binits{N.C.}},
\bauthor{\bsnm{Rotemberg}, \binits{V.}},
\bauthor{\bsnm{Halpern}, \binits{A.C.}},
\bauthor{\bsnm{Reiter}, \binits{O.}},
\bauthor{\bsnm{Carrera}, \binits{C.}},
\bauthor{\bsnm{Barreiro}, \binits{A.}},
\bauthor{\bsnm{Helba}, \binits{B.}}, \betal:
\batitle{Bcn20000: Dermoscopic lesions in the wild}.
\bjtitle{Scientific data}
\bvolume{11}(\bissue{1}),
\bfpage{641}
(\byear{2024})
\end{barticle}
\endbibitem

\bibitem[\protect\citeauthoryear{Codella et~al.}{2018}]{msk}
\begin{bchapter}
\bauthor{\bsnm{Codella}, \binits{N.C.}},
\bauthor{\bsnm{Gutman}, \binits{D.}},
\bauthor{\bsnm{Celebi}, \binits{M.E.}},
\bauthor{\bsnm{Helba}, \binits{B.}},
\bauthor{\bsnm{Marchetti}, \binits{M.A.}},
\bauthor{\bsnm{Dusza}, \binits{S.W.}},
\bauthor{\bsnm{Kalloo}, \binits{A.}},
\bauthor{\bsnm{Liopyris}, \binits{K.}},
\bauthor{\bsnm{Mishra}, \binits{N.}},
\bauthor{\bsnm{Kittler}, \binits{H.}}, \betal:
\bctitle{Skin lesion analysis toward melanoma detection: A challenge at the 2017 international symposium on biomedical imaging (isbi), hosted by the international skin imaging collaboration (isic)}.
In: \bbtitle{ISBI},
pp. \bfpage{168}--\blpage{172}
(\byear{2018})
\end{bchapter}
\endbibitem

\bibitem[\protect\citeauthoryear{Johnson et~al.}{2021}]{mimiciv}
\begin{botherref}
\oauthor{\bsnm{Johnson}, \binits{A.}},
\oauthor{\bsnm{Bulgarelli}, \binits{L.}},
\oauthor{\bsnm{Pollard}, \binits{T.}},
\oauthor{\bsnm{Celi}, \binits{L.A.}},
\oauthor{\bsnm{Mark}, \binits{R.}},
\oauthor{\bsnm{Horng~IV}, \binits{S.}}:
Mimic-iv-ed.
PhysioNet
(2021)
\end{botherref}
\endbibitem

\bibitem[\protect\citeauthoryear{Agarwal et~al.}{2018}]{dpd}
\begin{bchapter}
\bauthor{\bsnm{Agarwal}, \binits{A.}},
\bauthor{\bsnm{Beygelzimer}, \binits{A.}},
\bauthor{\bsnm{Dud{\'\i}k}, \binits{M.}},
\bauthor{\bsnm{Langford}, \binits{J.}},
\bauthor{\bsnm{Wallach}, \binits{H.}}:
\bctitle{A reductions approach to fair classification}.
In: \bbtitle{ICML},
pp. \bfpage{60}--\blpage{69}
(\byear{2018})
\end{bchapter}
\endbibitem

\bibitem[\protect\citeauthoryear{Agarwal et~al.}{2019}]{deodds}
\begin{bchapter}
\bauthor{\bsnm{Agarwal}, \binits{A.}},
\bauthor{\bsnm{Dud{\'\i}k}, \binits{M.}},
\bauthor{\bsnm{Wu}, \binits{Z.S.}}:
\bctitle{Fair regression: Quantitative definitions and reduction-based algorithms}.
In: \bbtitle{ICML},
pp. \bfpage{120}--\blpage{129}
(\byear{2019})
\end{bchapter}
\endbibitem

\end{thebibliography}

\end{document}